%% file: ai2_paper.tex
\documentclass[table]{ai2style/ai2}

\usepackage{microtype}
\usepackage{hyperref}
\usepackage{url}
\usepackage{booktabs, tabularx} 
\usepackage{graphicx}
\usepackage{lineno}
\usepackage{enumitem}
\usepackage{listings} 
\usepackage{svg}

\definecolor{ darkblue}{rgb}{0, 0, 0.5}
\hypersetup{colorlinks=true, citecolor=darkblue, linkcolor=darkblue, urlcolor=darkblue}

\usepackage{amssymb}
\usepackage{multirow}
\usepackage{bigdelim}
\usepackage{todonotes}
\usepackage{longtable}
\usepackage{tabularray}
\usepackage{wrapfig}
\usepackage[most]{tcolorbox}
\usepackage{url}
\usepackage{xspace}
\usepackage{svg}
\usepackage[absolute]{textpos} 

\usepackage{fdsymbol}   

\usepackage[utf8]{inputenc} 
\usepackage[T1]{fontenc}    
\usepackage{url}            
\usepackage{booktabs}       
\usepackage{amsfonts}       
\usepackage{nicefrac}       
\usepackage{microtype}      
\usepackage{amsmath}
\usepackage[most]{tcolorbox}
\usepackage{csquotes}

\usepackage{siunitx}
\usepackage{graphicx}
\usepackage{arydshln}
\usepackage{wrapfig}
\usepackage{enumitem}
\usepackage{soul} 

\usepackage{multirow}
\usepackage{xspace}
\usepackage{adjustbox}
\usepackage{pifont}
\usepackage{caption}
\usepackage{makecell}
\usepackage{subcaption}
\usepackage{bold-extra}
\usepackage{url}
\usepackage{xcolor,colortbl}

\usepackage{pgf-pie}

\usepackage{hyperref}
\definecolor{linkcolor}{RGB}{0, 0, 128}
\hypersetup{
     colorlinks   = true,
     citecolor    = linkcolor,
     linkcolor    = linkcolor,
     urlcolor     = linkcolor,
}
\usepackage{pifont}
\usepackage{listings}

\setlist[itemize]{leftmargin=*,itemsep=0em,parsep=0.3em,topsep=0.3em}

\DeclareUnicodeCharacter{2212}{\ensuremath{-}}

\usepackage{tikz}

\usepackage{setspace}

\usepackage{nicematrix}
\newcolumntype{L}[1]{>{\raggedright\let\newline\\\arraybackslash\hspace{0pt}}m{#1}}
\newcolumntype{C}[1]{>{\centering\let\newline\\\arraybackslash\hspace{0pt}}m{#1}}
\newcolumntype{R}[1]{>{\raggedleft\let\newline\\\arraybackslash\hspace{0pt}}m{#1}}
\newcolumntype{P}[1]{>{\centering\let\newline\\\arraybackslash\columncolor{ai2lightpink}}m{#1}}
\usepackage{listings}
\usepackage{xcolor}
\newcommand{\aitoo}{\raisebox{-1.5pt}{\includegraphics[height=1.05em]{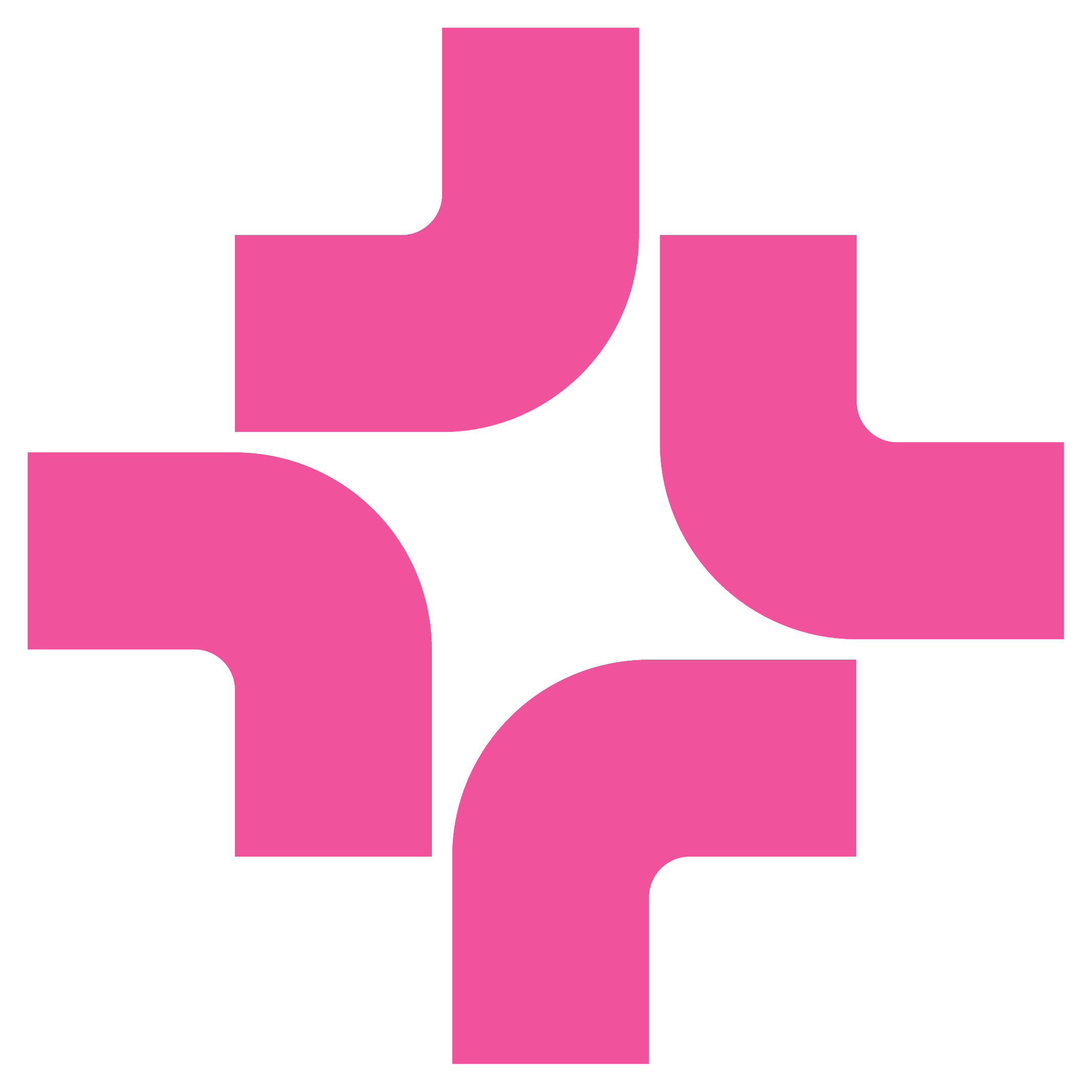}}\xspace}

\newcommand{\starOlmo}{\raisebox{.28em}{\hspace{.05em}\includegraphics[height=.5em]{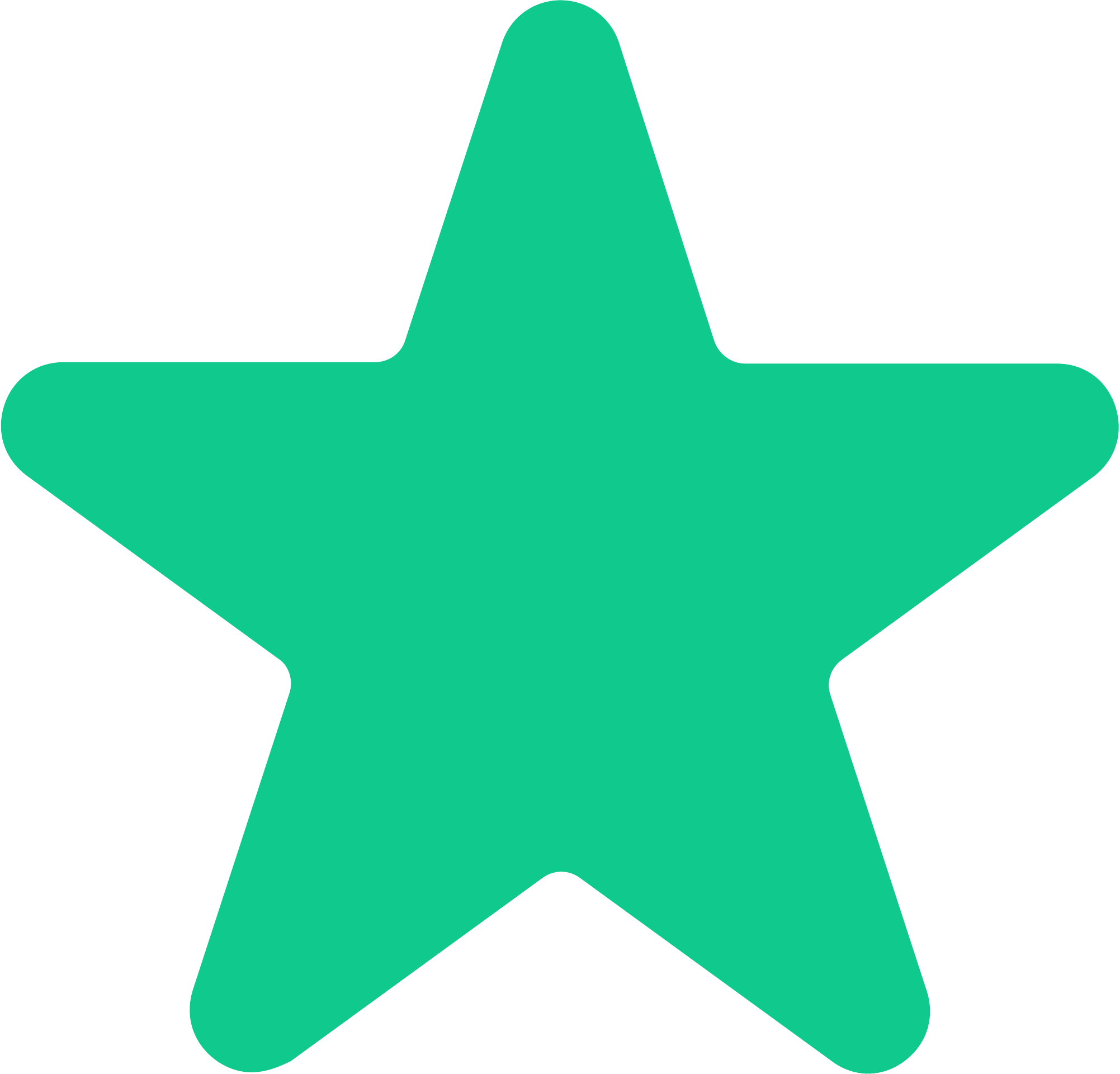}}}
\newcommand{\huggingface}{\raisebox{-1.5pt}{\includegraphics[height=1.05em]{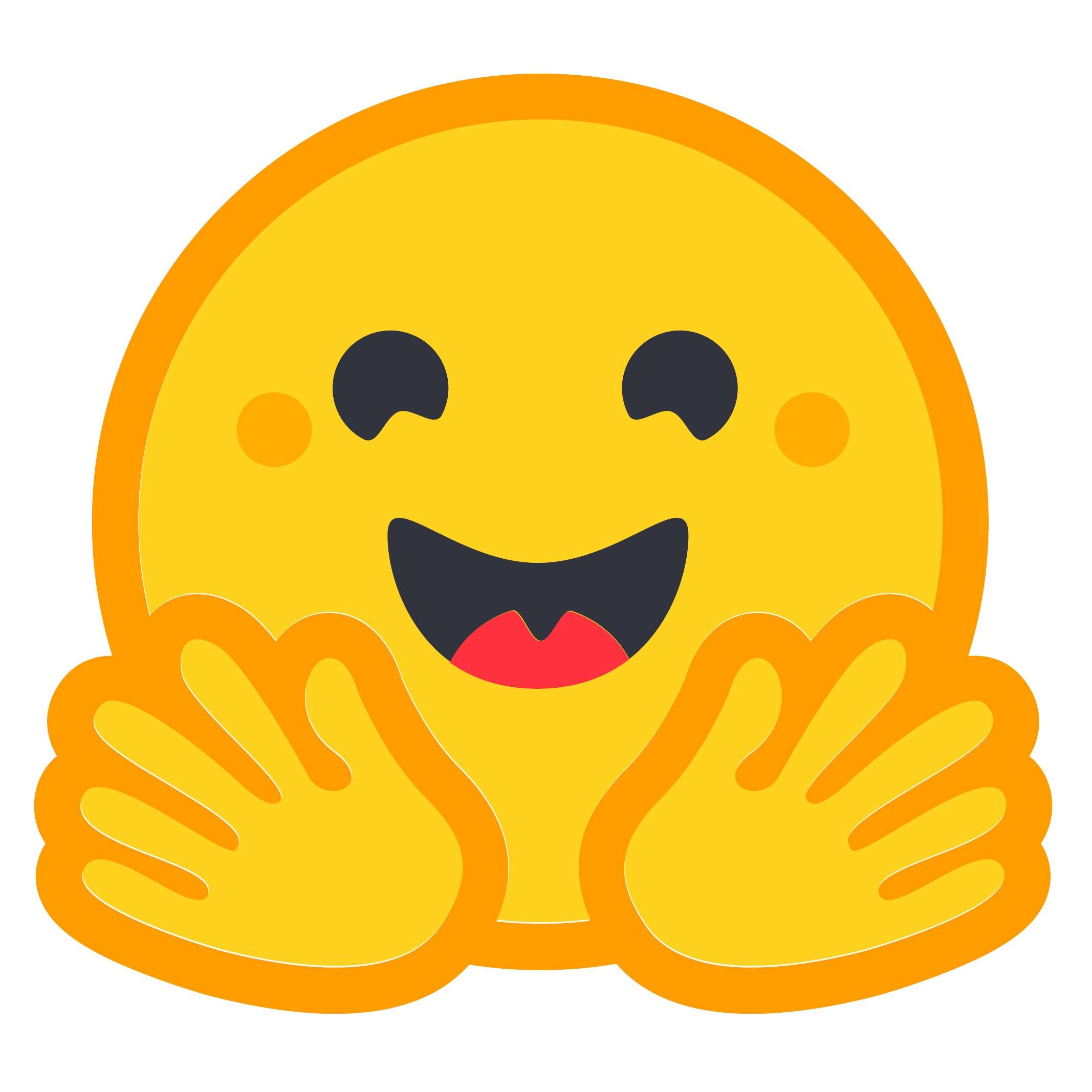}}\xspace}

\newcommand{\emailLogo}{\raisebox{-1.5pt}{\includegraphics[height=1.05em]{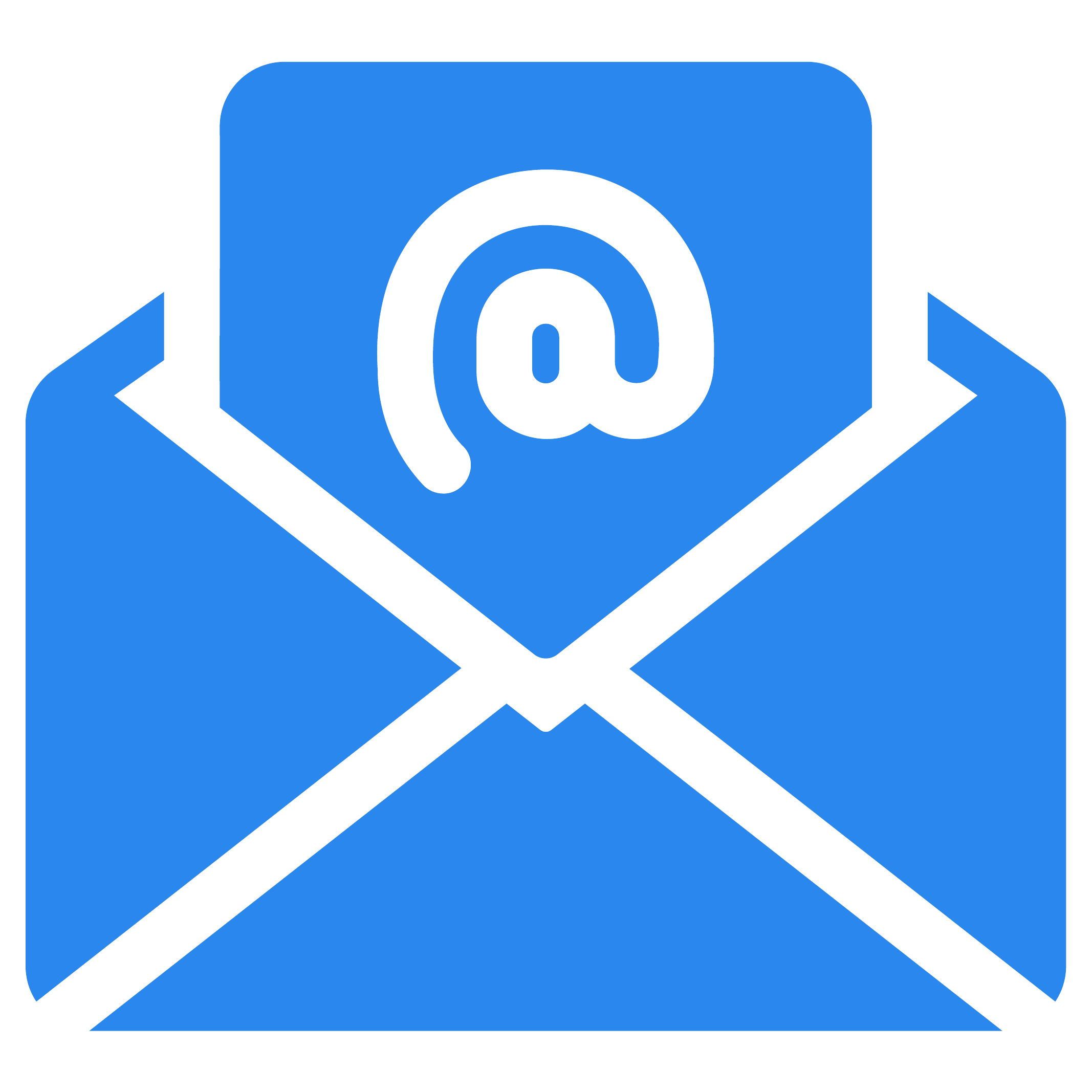}}\xspace}
\newcommand{\github}{\raisebox{-1.5pt}{\includegraphics[height=1.05em]{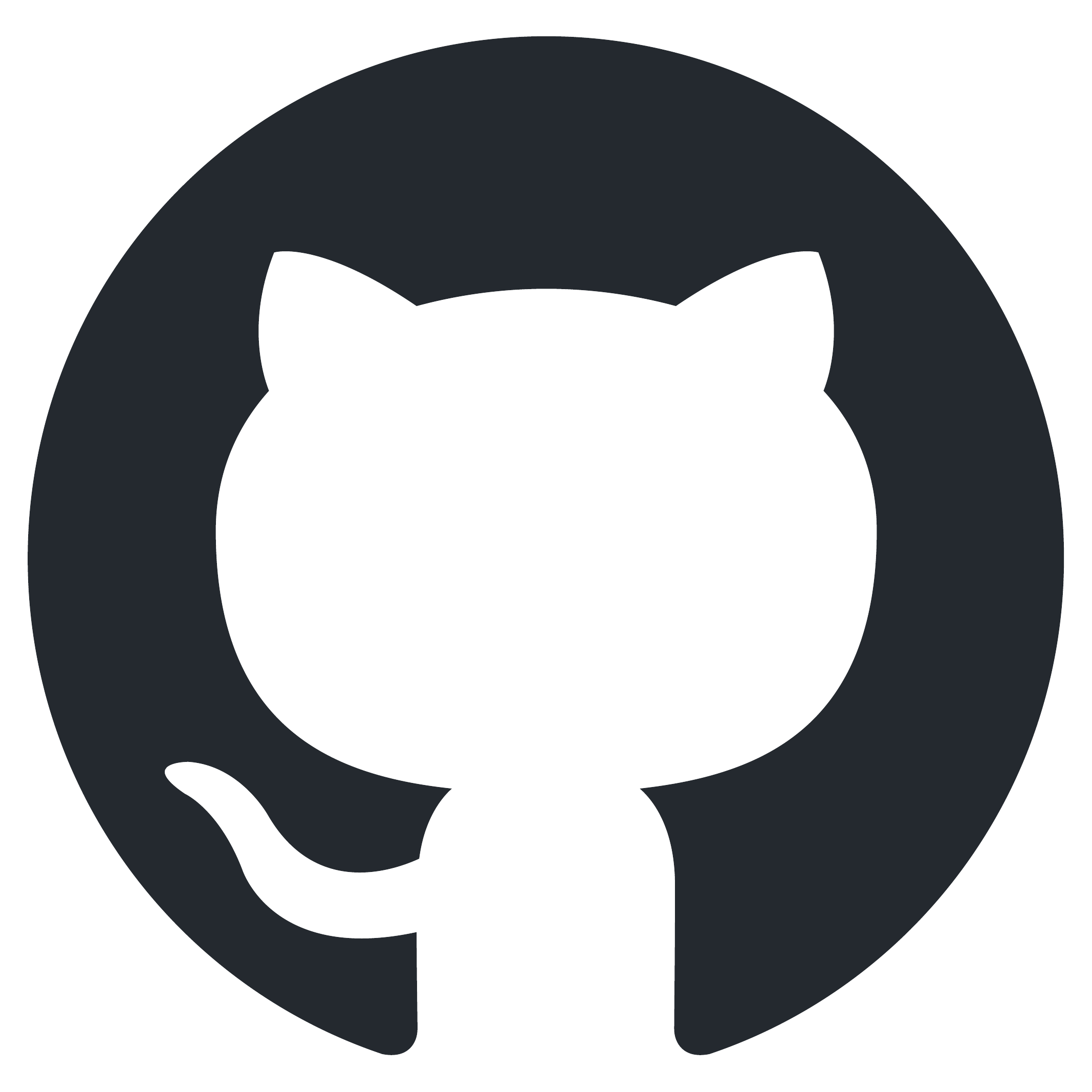}}\xspace}

\input{sec/defs}

\title{\model \space v1.\reflectbox{3}mbeddings}

\authorOne[\starOlmo]{Team OlmoEarth}
\authorTwo{Gabriel Tseng}
\authorTwo{Favyen Bastani}
\authorTwo{Patrick Alan Johnson}
\authorTwo{Yawen Zhang}
\authorTwo{Piper Wolters}
\authorThree{Ando Shah}
\authorThree{Hadrien Sablon}
\authorThree{Joseph Redmon}
\authorThree{Heather DeHaven}
\authorThree{Rachel Ratner}
\authorFour{Sasha Levy}
\authorFour{Jordan Steward}
\authorFour{Katrina Midgley}
\authorFour{Andrew Howe}
\authorFour{Hunter Pitelka}
\authorFour{Ted Schmitt}
\affiliation{Allen Institute for AI}
\contribution[]{\starOlmo {\model} is a team effort.}

\abstract{\input{sec/0_abstract}}

\metadata[\quad\aitoo Embeddings:]{\href{https://olmoearth.allenai.org/embeddings}{\texttt{olmoearth.allenai.org/embeddings}}}

\metadata[\vspace{.5em}\quad\github Training Code:]{
    \href{https://github.com/allenai/olmoearth_pretrain}{\texttt{olmoearth\_pretrain}}
}

\metadata[\vspace{.5em}\quad\huggingface OlmoEarth Pre-trained Model:]{
    \href{https://huggingface.co/allenai/OlmoEarth-v1_3-Base}{\texttt{OlmoEarth-v1\_3-Base}}
}

\metadata[\vspace{.5em}\quad\huggingface OlmoEarth Pre-training Dataset:]{
\href{https://huggingface.co/datasets/allenai/olmoearth_pretrain_dataset}{\texttt{olmoearth\_pretrain\_dataset}}}

\metadata[\vspace{.5em}\quad\emailLogo Contact:]{\color{ai2accent}{\texttt{olmoearth@allenai.org}}}

\begin{document}

\definecolor{colordefault}{HTML}{\colordefault}
\definecolor{colorgrey}{HTML}{\colorgrey}
\definecolor{colorone}{HTML}{\colorone}
\definecolor{colortwo}{HTML}{\colortwo}
\definecolor{colorthree}{HTML}{\colorthree}
\definecolor{colorfour}{HTML}{\colorfour}
\definecolor{colorfive}{HTML}{\colorfive}
\definecolor{colorsix}{HTML}{\colorsix}
\maketitle
\input{sec/1_intro}
\input{sec/3_model}
\input{sec/4_experiments}
\input{sec/5_platform}
\input{sec/5a_env_impact}
\input{sec/6_discussion}

\clearpage
\bibliographystyle{ieeenat_fullname}
\bibliography{main}

\clearpage

\input{sec/appendix}

\end{document}

%% file: sec/defs.tex
\usepackage{etoolbox}
\newtoggle{anon}
\togglefalse{anon}

\def\model{OlmoEarth}

\def\colordefault{0a3235}
\def\colorone{f0529c}
\def\colortwo{b11be8}
\def\colorthree{6fa8dc}
\def\colorfour{0fcb8c}
\def\colorfive{12cce5}
\def\colorsix{f65834}
\def\colorgrey{B0B0B0}

\usepackage{multirow}
\usepackage{comment}
\usepackage{adjustbox}

\usepackage{tabularx}
\usepackage{graphicx}
\usepackage{tikz}
\usetikzlibrary{patterns.meta,positioning, arrows.meta,fit,calc,shapes.geometric,matrix}

\usepackage{pgfplots}
\pgfplotsset{compat=1.18}

\newcommand*{\hdr}[1]{\multicolumn{1}{l}{\rlap{\,\,\,\rotatebox[origin = lb]{45}{\textbf{\scriptsize #1}}}}}

\usepackage{pdfrender}

\usepackage{graphicx}
\makeatletter
\def\maxwidth#1{\ifdim\Gin@nat@width>#1 #1\else\Gin@nat@width\fi}
\makeatother

\usepackage{contour}
\contourlength{1pt} 
\contournumber{20}  

\usepackage{pifont}

\def\ccheck{\color{black}{\ding{52}}}
\def\ccross{\color{black}{\ding{56}}}

%% file: sec/0_abstract.tex
We make some {\model} embeddings. To do this, we train a new model: OlmoEarth v1.3. OlmoEarth v1.3 is designed to condense multimodal, multitemporal remote sensing data into a single digestible 64 or 128-dimensional vector. OlmoEarth v1.3 produces state-of-the-art embeddings. In addition, OlmoEarth v1.3 is a strong encoder backbone -- competitive with OlmoEarth v1 and v1.2 -- while requiring 18\% and 53\% of the GPU hours to train, respectively. We use OlmoEarth v1.3 to compute global, annualized 10m embeddings, which we release under the OlmoEarth artifact license.

%% file: sec/1_intro.tex
\section{Introduction}
\label{sec:intro}

\input{sec/bytes_vs_rank}

Embeddings are an extremely convenient way to represent Earth Observations. When making large scale maps, inference costs dominate: Section 4 of the OlmoEarth v1.2 report \cite{olmoearth1_2} notes that when making maps at scale with OlmoEarth v1.2, 98\% of the cost is inference through the model and 2\% of the cost is data export \& preprocessing. Embeddings bear the inference cost once but can be reused for many downstream tasks. In addition, embeddings enable use cases not possible when finetuning models, such as aggregating information over large scales \cite{ma2026harvesting}. 

\footnotetext{For the AEF evaluations, we rank models using a single classifier. We use $k$NN-20 for consistency with Section \ref{sec:knnlp}.}

Previous generations of OlmoEarth models (OlmoEarth v1 \cite{olmoearth} and v1.2 \cite{olmoearth1_2}) have principally been deployed via finetuning within OlmoEarth Studio (\url{https://olmoearth.allenai.org/platform}). This tech report introduces OlmoEarth v1.3, a model which makes embedding a first class deployment target. Embeddings introduce three important considerations: (i) \textbf{aggregation}: combining tokens from multiple timesteps and modalities into a single representation per location, (ii) \textbf{efficiency} at high resolutions and (iii) \textbf{compaction}: decoupling the width of the backbone from the final embedding size.

OlmoEarth v1.3 yields state-of-the-art embeddings. When compared with the other two prominent 10m satellite embedding products -- Alpha Earth Foundations \cite{brown2025alphaearthfoundationsembeddingfield} and TESSERA v2 (specifically TESSERA-v2-Large) \cite{feng2026tessera} -- OlmoEarth v1.3's 128-dimensional embeddings are best on 10/11 tasks\footnote{As in Figures \ref{fig:result_summary_embeddings} and \ref{fig:result_summary}, we only consider the $k$NN-20 classifier for this summary.}, and our 64-dimensional embeddings beat both baselines on 7/11 tasks. In addition, OlmoEarth v1.3 is a strong finetuning model, maintaining OlmoEarth v1.2's quality despite requiring about half the GPU hours to train. 

In Section \ref{sec:model} we describe OlmoEarth v1.3, focusing on how we address the aggregation, efficiency and compaction considerations necessary to make strong global-scale embeddings. In Section \ref{sec:experiments}, we evaluate OlmoEarth v1.3's embeddings, and the backbone itself using the same $k$NN/LP and finetuning evals used in OlmoEarth v1.2. In Section \ref{sec:platform}, we describe our process for generating and serving global, multi-year embeddings from OlmoEarth v1.3. In Section \ref{sec:env_impact}, we discuss the environmental impact of pretraining OlmoEarth v1.3 and of deploying global, annualized embeddings.

%% file: sec/bytes_vs_rank.tex
\begin{wrapfigure}{r}{0.5\linewidth}
    \vspace{-8pt}
    \centering
    \begin{tikzpicture}
        \tikzset{modelmarker/.style={circle, inner sep=1.6pt, draw=black}}
        \begin{axis}[
            width=\linewidth,
            height=4.6cm,
            clip=false,
            xmin=40, xmax=156,
            ymin=0.6, ymax=3.8,
            xtick={64, 128},
            ytick={1, 2, 3},
            xlabel={Bytes per pixel},
            ylabel={Mean rank},
            xlabel style={font=\scriptsize, yshift=3pt},
            ylabel style={font=\scriptsize, yshift=-4pt},
            tick label style={font=\scriptsize},
            grid=major,
            y dir=reverse,
        ]
        \coordinate (aef)  at (axis cs:64,  3.50);
        \coordinate (d64)  at (axis cs:64,  2.32);
        \coordinate (d128) at (axis cs:128, 1.09);
        \coordinate (tes)  at (axis cs:132, 3.09);

        \draw[colorone,-] (d64) -- (d128);

        \node[modelmarker, fill=colortwo] at (aef) {};
        \node[anchor=west, text=colortwo] at (aef) {\scriptsize AEF};

        \node[modelmarker, fill=colorthree] at (tes) {};
        \node[anchor=north, text=colorthree, yshift=-1pt] at (tes) {\scriptsize TESSERA v2};

        \node[modelmarker, fill=colorone] at (d64) {};
        \node[modelmarker, fill=colorone] at (d128) {};
        \node[anchor=south east, text=colorone, xshift=2pt] at (d128) {\scriptsize {\model} v1.3};
        \node[anchor=north, text=colorone] at (d64)  {\tiny d64};
        \node[anchor=north, text=colorone] at (d128) {\tiny d128};
        \end{axis}
    \end{tikzpicture}
    \vspace{-6pt}
    \caption{Storage cost per pixel against mean rank (the smaller the better) over 11 evaluations (Section \ref{sec:embedding_results})\protect\footnotemark, all four products ranked together. TESSERA v2 stores 128 int8 values plus a float scale.}
    \label{fig:result_summary_embeddings}
    \vspace{-8pt}
\end{wrapfigure}
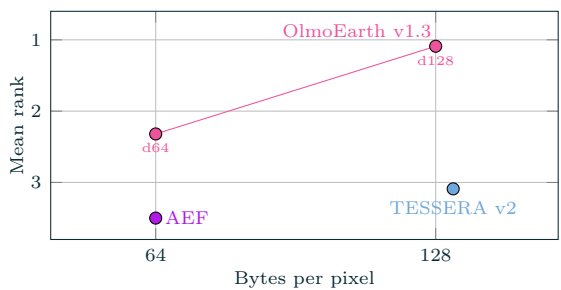

%% file: sec/3_model.tex
\section{OlmoEarth v1.3}
\label{sec:model}

\input{sec/v13_overview_figure}

The OlmoEarth models (v1 \cite{olmoearth}, v1.1, v1.2 \cite{olmoearth1_2} and v1.3) are encoder-decoder vision-transformer models trained via masked image modeling. The models are trained on multimodal, multitemporal remote sensing inputs. During training, part of the input is masked, and the model learns to reconstruct the masked part and other map targets based on the unmasked portion. 

Previous generations of OlmoEarth models were principally deployed within OlmoEarth Studio via finetuning; v1.3 makes embeddings a first-class deployment target. Unlike finetuning, three considerations are introduced when producing embeddings: (i) \textbf{aggregation}, (ii ) \textbf{efficiency} at high resolutions and (ii) \textbf{compaction}. 

Below, we describe how we handle these considerations. These represent the changes between OlmoEarth v1.2 and v1.3. We finish by summarizing the overall OlmoEarth v1.3 architecture and training loop. 

\subsection{Aggregation}\label{sec:agg}

\input{sec/perceiver_diagram}

The OlmoEarth models are based on vision transformers. Given an input of remote sensing data, we tokenize it before passing it through a number of attention blocks. This tokenization happens across three dimensions: space, time and modalities. Spatially, we define some patch size $p$ and divide our inputs $H \times W$ into $p \times p$ patches. A token consists of one modality, one timestep and one $p\times p$ patch so that a $T$-timestep, 3 modality (Sentinel-1, Sentinel-2 and Landsat) input will have $T \times 3 \times \frac{H}{p} \times \frac{W}{p}$ tokens in total. When constructing embeddings, we only want a single token per spatial patch, so we must aggregate the modality and time dimensions to produce $\frac{H}{p} \times \frac{W}{p}$ tokens. 

Naively averaging these embeddings can lead to detrimental results - for example, OlmoEarth v1.2 sees mixed results when combining modalities, even when finetuned (Table 2 of \cite{olmoearth1_2}). This makes intuitive sense; nothing during training tells the model that tokens will be averaged in this way. In addition, naive averaging is probably not optimal anyway; some tokens matter much more than others. For example, during cloudy months Sentinel-1 is likely to be more important.

We aggregate the encoder's output tokens using an Attention Pooler (a relatively common approach \cite{jaegle2022perceiver,lee2019set,jaegle2021perceiver,li2023blip}). Specifically, we learn a query and the ``tokens to be aggregated'' are the keys and values. In our case, the key / value tokens are the outputs of the vision transformer. This allows the vision transformer's output tokens to be aggregated using a data dependent aggregation function -- the attention mechanism. Our Attention Pooler consists of two attention layers; a ``read'' cross-attention layer, and a latent self-attention layer (where the queries, keys and values are all the query tokens). During the reads, the keys and values are the final outputs of the Vision Transformer; we experimented with passing intermediate embeddings from the Vision Transformer too but this did not meaningfully change results.

We have two requirements for these query tokens: (i) locality -- each query token must correspond to a specific spatial patch, and (ii) window invariance -- windowing is arbitrary, so nothing should depend on a token's absolute position within the window. We achieve this with three mechanisms:

\textbf{A single learned location query} which is repeated to cover all spatial patches. This ensures window invariance, since every spatial patch shares the same query parameters. We refer to the location queries after they have passed through the Attention Pooler as ``location tokens''. 

\textbf{RoPE spatial encodings}, applied during both the read and self-attention layers, give each query its spatial identity. Because RoPE is a \textit{relative} positional encoding, it allows us to provide this spatial identity without imposing a dependency on a query token's absolute position within a window.

\textbf{Direct supervision on location tokens}. This direct supervision leverages our map modalities: we add a loss on each query token, training it to reconstruct the map pixels it covers spatially. Concretely, each location token covers a $p \times p$ spatial area in the input. For each map modality, we learn a linear projection (which we discard after pretraining) which maps each 768-dimensional output token to $P \times P \times n$ where $P$ is our maximum location token patch size and $n$ is the number of channels in that map. We then bilinearly resize to the token's footprint and apply an L1 loss for continuous maps (like SRTM) and a cross-entropy loss for categorical maps (like WorldCover). The full supervision loss is a weighted sum of each map $m$'s loss:

$$
\mathcal{L}_{\mathrm{sup}} = \sum_{m} w_m  \mathcal{L}_m
$$

For the continuous maps, a mean prediction incurs a loss of $\mathbb{E}|x-\mu|\le\sigma$; under our normalization, $\sigma \approx 0.25$. For the categorical maps, a uniform prediction incurs a loss of $\ln C$, which is 2.4 for WorldCover's 11 classes and 4.9 for CDL's 133 classes. We therefore apply a weight $w_m = 1$ for the continuous maps and $w_m = 0.1$ for categorical maps to balance out these order-of-magnitude differences. 

Direct supervision improves embedding performance; PASTIS validation mIoU increases from $51.3$ without supervision to $58.4$ with supervision (we describe how we run PASTIS in Section \ref{sec:embedding_results}). 

\subsection{Efficiency at high resolutions}

\input{sec/macs_per_patch_size}

All OlmoEarth models use a flexible patch-embedding layer; this allows users to control the inference cost of their models (Figure \ref{fig:patch_size_macs}). This patch size defines the resolution at which an input is tokenized; given a 10m per pixel input, a patch size of 2 means that each token represents a 20m $\times$ 20m spatial patch. In a standard ViT, the input resolution is equal to the output resolution; that is, at patch size 2 the ViT will also \textit{output} tokens covering a 20m $\times$ 20m spatial patch.

During finetuning, this is not an issue; we can use a decoder head which upsamples these tokens to predictions at the desired resolution (typically 10m per pixel). However, when producing embeddings we don't have a decoder head - we'd like to decouple the encoder patch size (which dictates the inference compute cost) from the model's output resolution. The Attention Pooler provides a simple way to do this: we can repeat the learned queries for every desired output patch, instead of for every encoder spatial patch (Figure \ref{fig:perceiver}).

This allows us to produce 10m embeddings much more cheaply; Figure \ref{fig:patch_size_macs} shows the cost of adding an Attention Pooler to output 10m embeddings when running the ViT at different resolutions. The Attention Pooler adds a negligible cost (18 to 4 GMACs per window) while allowing significantly cheaper forward passes. 

Decoupling the ViT patch size from Attention Pooler's output resolution allows us to efficiently produce 10m embeddings - running a ViT + Attention Pooler at encoder resolution 20m, output resolution 10m is \textbf{7.7$\times$ cheaper} than running a ViT at 10m resolution: 308 vs. 2,378 GMACs per window (for reference, TESSERA v2 costs around 700-1,200 GMACs for the same area, depending on imagery availability). Figure \ref{fig:patch_size_macs} shows the overhead of the Attention Pooler outputting at 10m resolution at different input patch sizes. The Attention Pooler introduces a negligible additional cost to the ViT.

We already sample encoder patch sizes $p_e$ during training (Section 2.2 of \cite{olmoearth}). We now additionally sample pooling patch sizes $p_p$ such that $p_p \leq p_e$ - this ensures the pooler can output location tokens at different resolutions. We want the Attention Pooler's location tokens to retain fine-grained details. We use two mechanisms to encourage this; \textbf{direct supervision} on the output tokens, as described in Section \ref{sec:agg}, and \textbf{PixLatentMIM Lite}. This is a variation of our LatentMIM Lite loss (Section 2.4 of OlmoEarth v1, \cite{olmoearth}). The targets (and therefore the decoder's mask tokens) always operate at a pixel resolution (patch size 1). To prevent memory issues from too many decode-tokens, we sample one pixel per masked patch to use as a target. This yields a small improvement over standard LatentMIM Lite (+0.5 mIoU on the PASTIS validation set). 

\subsection{Compaction}\label{sec:compaction}

\input{sec/distillation}

A ViT-Base has 768-dimensional encodings; we would like to decouple the final output size from the model's capacity, so that much more compact embeddings can be produced. We tried two ways to achieve this: \textbf{query compaction}, where the Attention Pooler's queries are set at a desired dimensionality (e.g. 128), and \textbf{distillation}, where the entire backbone is kept at 768 dimensions before being shrunk via distillation.

Both of these approaches yielded similar performance on the 128d embeddings (distillation obtains an mIoU of $58.4$ on the PASTIS validation set while token compaction obtains $57.5$). Our final recipe uses distillation; we chose this because distillation offers two advantages over query compaction:

\begin{itemize}
\item \textbf{A strong teacher model}: Using 128 dimensional query tokens in the Attention Pooler forces a bottleneck through the Attention Pooler. When using distillation, the backbone is trained natively at d768. This ensures we have a strong teacher model, which also excels at finetuning. 
\item \textbf{Nested embeddings} via a Matryoshka learning \cite{NEURIPS2022_c32319f4}, allowing users to select d128 or d64 embeddings based on their own performance vs. cost tradeoffs. 
\end{itemize}

Our ``distillation'' consists of learning a single linear layer, which projects the 768 location tokens down to 128 dimensions, followed by a LayerNorm (Figure \ref{fig:distillation}). To train this linear layer, we use two losses: a cosine similarity loss ($\mathcal{L}_{\textrm{cosine}}$) and a Gram loss ($\mathcal{L}_{\textrm{Gram}}$, following DINOv3 \cite{siméoni2025dinov3}). The cosine similarity loss uses a 2-layer MLP with hidden width 256 to project the student embeddings back to 768 dimensions, before measuring their cosine similarity with the location tokens. The Gram loss randomly samples 2,048 tokens from the batch and constructs two cosine-similarity matrices for these tokens: one using the 768 dimensional location tokens, and one using the 128 dimensional student embeddings. An MSE loss is then applied to compare the two matrices. Unlike DINOv3's Gram loss, this means similarities are also computed between tokens in different instances. Intuitively, the cosine loss encourages effective distillation at a token level, while the Gram loss encourages the student to preserve the overall geometry of the tokens by matching the teacher's pairwise similarities between tokens. 

These distillation losses are \textbf{only used to train the linear projection} and the MLPs (which we discard after pretraining) -- gradients from the distillation losses do not flow to the ViT backbone or the Attention Pooler. We use a Matryoshka loss for the distillation; this means that the distillation losses are applied twice, once to the first 64 dimensions of the student tokens and once to the full 128 dimensions (we learn one MLP per prefix for the cosine loss). This allows us to treat the first 64 dimensions of the student tokens as an even smaller set of embeddings. 

We train our student jointly with the teacher (the ViT backbone and the Attention Pooler) rather than in a separate stage. Since the student only consists of a linear layer, this makes the distillation essentially free; a single forward pass allows us to compute the LatentMIM Lite and supervision losses for the teacher as well as distillation losses for the student.

\subsection{Putting it together}

OlmoEarth v1.3 follows the same encoder-decoder training recipe and architecture as OlmoEarth v1.2, but adds two components: the Attention Pooler (Section \ref{sec:agg}) and token distillation (Section \ref{sec:compaction}). We use a ViT-Base for our Vision Transformer; this ViT-Base is identical to the OlmoEarth v1.2 Base model. The decoder inputs when training OlmoEarth v1.3 are the attention pooler outputs, not the raw outputs from the Vision Transformer backbone. We provide a visual overview to OlmoEarth v1.3 in Figure \ref{fig:v1_3_overview}. 

All losses are simply summed during training:

$$
\mathcal{L} = \mathcal{L}_{\textrm{PixLatentMIM}} + \mathcal{L}_{\textrm{sup}} + \mathcal{L}_{\textrm{cosine}} + \mathcal{L}_{\textrm{Gram}}
$$

Where $\mathcal{L}_{\textrm{LatentMIM}}$ is the token-contrastive loss we used to reconstruct masked Sentinel-1, Sentinel-2 and Landsat tokens and the maps, $\mathcal{L}_{\textrm{sup}}$ is the supervision loss on maps, and $\mathcal{L}_{\textrm{cosine}}$ and $\mathcal{L}_{\textrm{Gram}}$ are cosine similarity and Gram loss used for distillation. 

We use the same training data as OlmoEarth v1.2:

\begin{table}[h]
\centering
    \begin{tabular}{l|ll}
    Observations & Maps & \\
    \hline
    Sentinel-1 & WorldCereal \cite{van2023worldcereal} & OpenStreetMap \cite{OpenStreetMap} \\
    Sentinel-2 & WorldCover \cite{zanaga2022esa} & Cropland Data Layer \cite{USDA_NASS_Cropland_Data_Layer}\\
    Landsat-8 & SRTM \cite{SRTM} & Canopy Height Map \cite{tolan2024very}\\
    \end{tabular}
\end{table}

One of the main motivations for the ``Instance Contrastive Loss'' (\cite{olmoearth}, Section 2.4.2) is to pool tokens during pre-training so that when we pool tokens during deployment this is something the model has seen before. The Attention Pooler now aggregates tokens both during pre-training and deployment, allowing us to remove the instance contrastive loss without seeing a regression in performance. This speeds training up, since gradient updates now only require one forward pass per microbatch instead of two. 

Aside from these changes, we train OlmoEarth v1.3 using the same recipe as OlmoEarth v1 and v1.2. This consists of using an AdamW optimizer with a batch size of 512 for 667,200 steps. We use a linear warmup of 8,000 steps and then apply a cosine annealing learning rate for the remainder of training. \textbf{OlmoEarth v1.3 cost $53\%$ of the GPU hours required to train OlmoEarth v1.2, and $18\%$ required to train OlmoEarth v1}. 

%% file: sec/v13_overview_figure.tex
\definecolor{oeink}{HTML}{0A3235}
\definecolor{oeteal}{HTML}{105257}
\definecolor{oecream}{HTML}{FAF2E9}
\definecolor{oepink}{HTML}{F0529C}
\definecolor{oepurple}{HTML}{B11BE8}
\definecolor{oegreen}{HTML}{0FCB8C}
\colorlet{oetokfill}{oeteal!12}
\colorlet{oepinkfill}{oepink!12}
\colorlet{oelossfill}{oeteal!10}
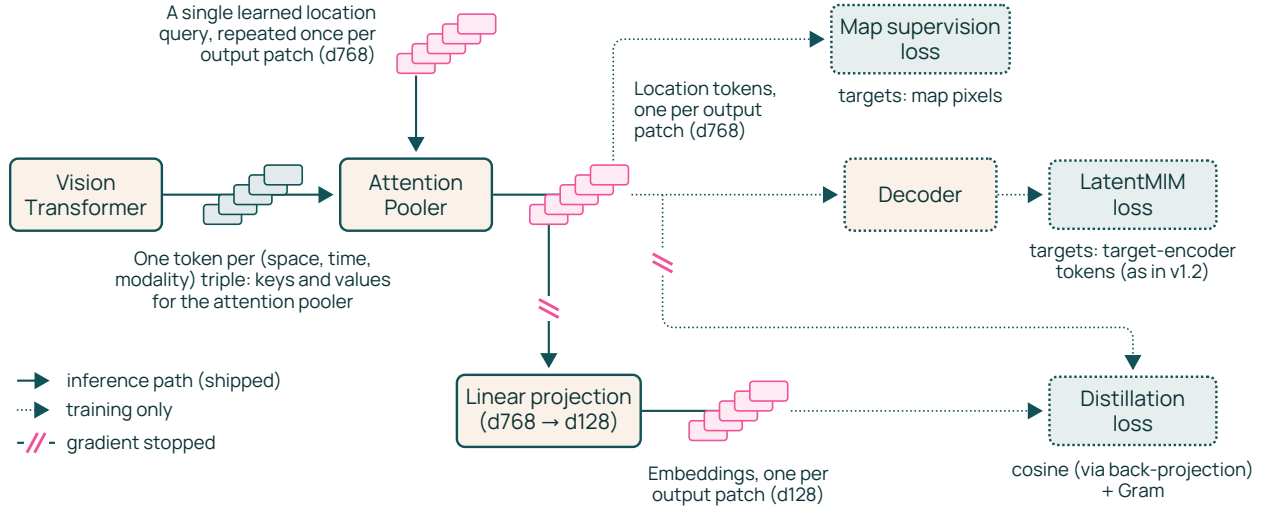
\begin{figure}
\centering
\resizebox{\linewidth}{!}{%
\begin{tikzpicture}[
    font=\sffamily\footnotesize,
    module/.style={draw=oeteal, line width=0.9pt, fill=oecream, rounded corners=3pt,
                   minimum height=0.95cm, minimum width=2.0cm, align=center, text=oeink},
    trainmodule/.style={module, draw=oeteal, densely dotted},
    loss/.style={draw=oeteal, densely dotted, line width=0.9pt, fill=oelossfill, rounded corners=3pt,
                 minimum height=0.9cm, minimum width=2.2cm, align=center, text=oeink},
    note/.style={align=center, font=\sffamily\scriptsize, text=oeink},
    target/.style={align=center, font=\sffamily\scriptsize, text=oeink},
    arr/.style={-{Triangle[length=2.0mm, width=2.2mm]}, line width=0.8pt, draw=oeteal, shorten >=2pt},
    tarr/.style={arr, line width=0.6pt, densely dotted, rounded corners=4pt},
    tokens/.pic={
      \foreach \i in {0,...,4}
        \draw[fill=\tokfill, draw=\tokdraw, line width=0.6pt, rounded corners=1.2pt]
          (\i*0.2, \i*0.13) rectangle ++(0.5, 0.26);
      \coordinate (-w)  at (-0.04, 0.39);
      \coordinate (-e)  at (1.34, 0.39);
      \coordinate (-n)  at (0.65, 0.82);
      \coordinate (-s)  at (0.65, -0.04);
      \coordinate (-se) at (1.05, -0.02);
      \coordinate (-sw) at (-0.04, 0.13);
      \coordinate (-ne) at (1.34, 0.65);
    },
    teal tokens/.style={/utils/exec={\def\tokfill{oetokfill}\def\tokdraw{oeteal}}},
    pink tokens/.style={/utils/exec={\def\tokfill{oepinkfill}\def\tokdraw{oepink}}},
    sg/.pic={
      \fill[white] (-0.16, -0.15) rectangle (0.16, 0.15);
      \draw[oepink, line width=1.1pt] (-0.12, -0.13) -- (0.02, 0.13);
      \draw[oepink, line width=1.1pt] (-0.02, -0.13) -- (0.12, 0.13);
    },
  ]


  \node[module] (vit) at (0, 1.15) {Vision\\Transformer};
  \node[module] (perc) at (4.35, 1.15) {Attention\\Pooler};
  \node[module, minimum width=2.3cm] (proj) at (6.1, -1.7)
    {Linear projection\\(d768 $\to$ d128)};

  \draw[arr] (vit.east) -- (perc.west);
  \pic[teal tokens] (kv) at (1.55, 0.76) {tokens};

  \draw[line width=0.8pt, draw=oeteal] (perc.east) -- (6.6, 1.15);
  \draw[arr] (6.1, 0.9) -- (proj.north);
  \pic[pink tokens] (out) at (5.85, 0.76) {tokens};
  \pic[rotate=90] at (6.1, -0.35) {sg};

  \draw[line width=0.8pt, draw=oeteal] (proj.east) -- (8.6, -1.7);
  \pic[pink tokens] (emb) at (7.95, -2.09) {tokens};

  \pic[pink tokens] (q) at (4.1, 2.75) {tokens};
  \draw[arr] (4.35, 2.72) -- (perc.north);

  \node[loss] (mapsup) at (11.0, 3.2) {Map supervision\\loss};
  \node[target, below=0.06cm of mapsup.south] {targets: map pixels};
  \node[trainmodule] (dec) at (11.0, 1.15) {Decoder};
  \node[loss] (mim) at (13.8, 1.15) {LatentMIM\\loss};
  \node[target, below=0.06cm of mim.south] {targets: target-encoder\\tokens (as in v1.2)};
  \node[loss] (distill) at (13.8, -1.7) {Distillation\\loss};
  \node[target, below=0.06cm of distill.south]
    {cosine (via back-projection)\\$+$ Gram};

  \draw[tarr] (6.95, 1.6) |- (mapsup.west);
  \draw[tarr] (out-e |- 0, 1.15) -- (dec.west);
  \draw[tarr] (dec.east) -- (mim.west);
  \draw[tarr] (7.6, 1.15) -- (7.6, -0.6) -- (13.8, -0.6) -- (distill.north);
  \pic[rotate=90] at (7.6, 0.25) {sg};
  \draw[tarr] (emb-e |- 0, -1.7) -- (distill.west);

  \node[note, anchor=north] at (2.2, 0.55)
    {One token per (space, time,\\modality) triple: keys and values\\for the attention pooler};
  \node[note, anchor=east, align=right] at (3.95, 3.25)
    {A single learned location\\query, repeated once per\\output patch (d768)};
  \node[note, anchor=west, align=left] at (7.1, 2.25)
    {Location tokens,\\one per output\\patch (d768)};
  \node[note, anchor=north] at (8.6, -2.3)
    {Embeddings, one per\\output patch (d128)};

  \node[anchor=north west, align=left, font=\sffamily\scriptsize, text=oeink, inner sep=2pt]
    at (-1.0, -1.1)
    {\tikz[baseline=-0.6ex]{\draw[arr] (0,0)--(0.5,0);} \, inference path (shipped)\\[3pt]
     \tikz[baseline=-0.6ex]{\draw[tarr] (0,0)--(0.5,0);} \, training only\\[3pt]
     \tikz[baseline=-0.6ex]{\draw[oeteal, line width=0.8pt] (0,0)--(0.5,0); \pic at (0.25,0) {sg};} \, gradient stopped};
\end{tikzpicture}%
}
\caption{One OlmoEarth v1.3 training step. In a single forward pass, we compute the LatentMIM Lite and supervision losses (which are used to train the Attention Pooler and Vision Transformer) and the distillation losses (which are used to train the student). This makes distillation essentially free: no additional forward passes are necessary to compute the distillation losses.}
\label{fig:v1_3_overview}
\end{figure}

%% file: sec/perceiver_diagram.tex
\definecolor{oeink}{HTML}{0A3235}
\definecolor{oeteal}{HTML}{105257}
\definecolor{oecream}{HTML}{FAF2E9}
\definecolor{oepink}{HTML}{F0529C}
\definecolor{oepurple}{HTML}{B11BE8}
\definecolor{oegreen}{HTML}{0FCB8C}
\colorlet{oetokfill}{oeteal!12}
\colorlet{oepinkfill}{oepink!12}
\begin{figure}[t]
\centering
\resizebox{\linewidth}{!}{%
\begin{tikzpicture}[
    font=\sffamily\footnotesize,
    module/.style={draw=oeteal, line width=0.9pt, fill=oecream, rounded corners=3pt,
                   minimum height=1.1cm, minimum width=2.3cm, align=center, text=oeink},
    note/.style={align=left, font=\sffamily\scriptsize, text=oeink},
    arr/.style={-{Triangle[length=2.2mm, width=2.4mm]}, line width=0.9pt, draw=oeteal},
    tokcell/.style={draw=oeteal!85, fill=oetokfill, line width=0.6pt},
    pixcell/.style={draw=oeteal!45, fill=white, line width=0.4pt},
    qcell/.style={draw=oepink, fill=oepinkfill, line width=0.6pt},
    hicell/.style={draw=oepink, fill=oepink, line width=0.6pt},
  ]
  \def\px{0.3}
  \def\pt{0.6}

  \foreach \k in {2,1,0} {
    \begin{scope}[shift={(\k*0.16, \k*0.13)}]
      \foreach \i in {0,...,3} \foreach \j in {0,...,3}
        \draw[pixcell] (\i*\px, \j*\px) rectangle ++(\px, \px);
    \end{scope}
  }
  \fill[oeteal!22] (0, 0.6) rectangle ++(0.6, 0.6);
  \foreach \i in {0,...,3} \foreach \j in {0,...,3}
    \draw[pixcell, fill=none] (\i*\px, \j*\px) rectangle ++(\px, \px);
  \draw[oeteal, line width=0.5pt, densely dashed] (0.6, 0) -- ++(0, 1.2) (0, 0.6) -- ++(1.2, 0);
  \node[note, align=center, anchor=north] at (0.76, -0.12)
    {Input pixels (10\,m)\\$T$ timesteps $\times$ $M$ modalities};

  \draw[arr] (1.75, 0.75) -- (3.0, 0.75);
  \node[note, align=center, anchor=south] at (2.35, 0.82) {patch\\embedding};
  \coordinate (itok) at (3.2, 0);
  \foreach \k in {2,1,0} {
    \begin{scope}[shift={($(itok) + (\k*0.16, \k*0.13)$)}]
      \foreach \i in {0,1} \foreach \j in {0,1}
        \draw[tokcell] (\i*\pt, \j*\pt) rectangle ++(\pt, \pt);
    \end{scope}
  }
  \draw[tokcell, fill=oeteal!40] ($(itok) + (0, 0.6)$) rectangle ++(\pt, \pt);
  \node[note, align=center, anchor=north] at ($(itok) + (0.76, -0.12)$)
    {Tokens: one per 20\,m\\patch (patch size 2),\\timestep and modality};

  \node[module] (vit) at (6.45, 0.75) {Vision\\Transformer};
  \draw[arr] (4.85, 0.75) -- (vit.west);

  \begin{scope}[xshift=2.9cm]
  \draw[draw=oeteal, line width=0.9pt, fill=oecream!60, rounded corners=4pt]
    (5.3, -1.5) rectangle (10.8, 2.55);
  \node[text=oeink, font=\sffamily\small\bfseries, anchor=north west] at (5.45, 2.45)
    {Cross-attention};

  \coordinate (tok) at (6.35, 0.15);
  \foreach \k in {2,1,0} {
    \begin{scope}[shift={($(tok) + (-\k*0.16, \k*0.13)$)}]
      \foreach \i in {0,1} \foreach \j in {0,1}
        \draw[tokcell] (\i*\pt, \j*\pt) rectangle ++(\pt, \pt);
    \end{scope}
  }
  \draw[arr] (vit.east) -- ($(tok) + (-0.4, 0.5)$);

  \coordinate (qg) at (9.0, 0.15);
  \foreach \i in {0,...,3} \foreach \j in {0,...,3}
    \draw[qcell] ($(qg) + (\i*\px, \j*\px)$) rectangle ++(\px, \px);
  \draw[hicell] ($(qg) + (1*\px, 2*\px)$) rectangle ++(\px, \px);
  \draw[oeteal, line width=0.5pt, densely dashed]
    ($(qg) + (0.6, 0)$) -- ++(0, 1.2)  ($(qg) + (0, 0.6)$) -- ++(1.2, 0);

  \coordinate (hq) at ($(qg) + (1.5*\px, 2.5*\px)$);
  \foreach \i/\j/\w in {0/0/0.5, 1/0/0.9, 0/1/1.6, 1/1/1.1}
    \draw[oepink, line width=\w pt, opacity=0.85]
      (hq) -- ($(tok) + (\i*\pt + 0.5*\pt, \j*\pt + 0.5*\pt)$)
      node[circle, fill=oepink, inner sep=0pt, minimum size=3pt] {};
  \fill[oepink] (hq) circle (1.3pt);
  \node[note, align=center, anchor=south, text=oepink] at (8.3, 1.12)
    {attends to\\every token};

  \node[note, align=center, anchor=north] at ($(tok) + (0.45, -0.12)$)
    {\textbf{Keys, values}:\\ViT output tokens\\(20\,m)};
  \node[note, align=center, anchor=north] at ($(qg) + (0.6, -0.12)$)
    {\textbf{Queries}:\\one per 10\,m\\output pixel};

  \draw[qcell, rounded corners=1pt] (9.45, 3.2) rectangle ++(0.3, 0.3);
  \node[note, anchor=west] at (9.9, 3.35)
    {A single learned query,\\repeated once per output pixel};
  \draw[arr] (9.6, 3.18) -- ($(qg) + (0.6, 1.3)$);

  \node[module] (self) at (12.45, 0.75) {Self-attention\\(between queries)};
  \draw[arr] (10.8, 0.75) -- (self.west);

  \coordinate (og) at (14.55, 0.15);
  \foreach \i in {0,...,3} \foreach \j in {0,...,3}
    \draw[qcell] ($(og) + (\i*\px, \j*\px)$) rectangle ++(\px, \px);
  \draw[hicell] ($(og) + (1*\px, 2*\px)$) rectangle ++(\px, \px);
  \draw[arr] (self.east) -- ($(og) + (-0.1, 0.6)$);
  \node[note, align=center, anchor=north] at ($(og) + (0.6, -0.12)$)
    {Embeddings, one per\\output pixel (10\,m)};
  \end{scope}
\end{tikzpicture}%
}
\caption{The Attention Pooler learns a query which cross-attends to the ViT's output tokens. We repeat this query at the desired output resolution to obtain embeddings at a finer granularity than the ViT's patch size; this decouple the model's output resolution from the number of tokens passing through the ViT (and therefore its compute cost).}
\label{fig:perceiver}
\end{figure}
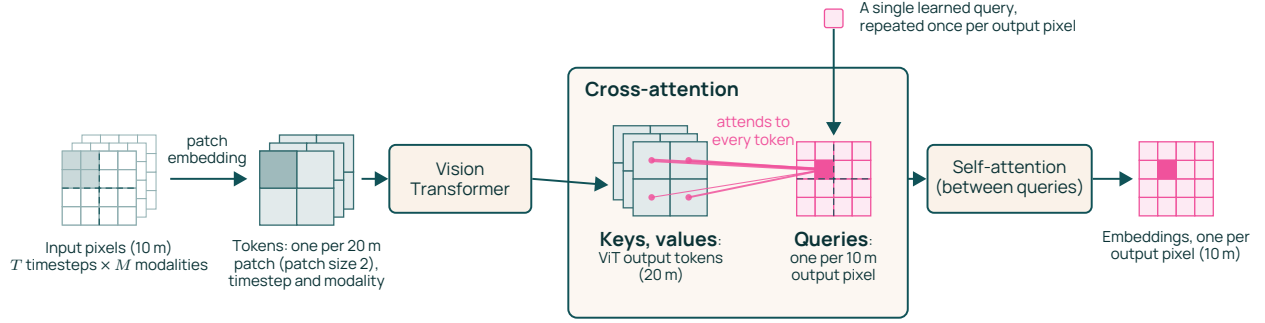

%% file: sec/macs_per_patch_size.tex
  \begin{wrapfigure}{r}{0.45\linewidth}
      \vspace{-8pt}
      \centering
      \begin{tikzpicture}
          \begin{axis}[
              width=\linewidth,
              height=4.2cm,
              clip=false,
              xmin=0.8, xmax=4.2,
              ymin=0, ymax=2600,
              xtick={1, 2, 4},
              xticklabels={1, 2, 4},
              ytick={0, 1000, 2000},
              yticklabels={0, 1000, 2000},
              xlabel={Patch size},
              ylabel={GMACs},
              xlabel style={font=\scriptsize, yshift=3pt},
              ylabel style={font=\scriptsize, yshift=-4pt},
              tick label style={font=\scriptsize},
              grid=major,
              legend style={font=\tiny, at={(0.5,1.03)}, anchor=south, inner sep=1pt, draw=none, column sep=4pt}, legend columns=2, legend image post style={xscale=0.6},
              legend cell align=left,
          ]
          \addplot[colordefault, mark=*, mark size=1.4pt, thick]
              coordinates {(1, 2377.6) (2, 300.9) (4, 56.9)};
          \addlegendentry{Encoder}
          \addplot[colorone, mark=square*, mark size=1.4pt, thick, dashed]
              coordinates {(1, 2395.5) (2, 307.9) (4, 61.2)};
          \addlegendentry{+ Attention Pooler}

          \node[anchor=west, text=colorone, font=\tiny, inner sep=1pt, xshift=2pt] at (axis cs:1, 2395.5) {\textcolor{colordefault}{2,378}+18};
          \node[anchor=south west, text=colorone, font=\tiny, inner sep=1pt, xshift=2pt] at (axis cs:2, 307.9) {\textcolor{colordefault}{301}+7};
          \node[anchor=south east, text=colorone, font=\tiny, inner sep=1pt, xshift=2pt] at (axis cs:4, 200) {\textcolor{colordefault}{57}+4};
          \end{axis}
      \end{tikzpicture}
      \vspace{-6pt}
      \caption{Inference cost of one $16 \times 16$ embedding window (12 timesteps, Sentinel-1, Sentinel-2 and Landsat) by patch size, with and without an Attention Pooler outputting at pixel-resolution. Labels give the encoder's GMACs and \textcolor{colorone}{the Pooler's overhead}.}
      \label{fig:patch_size_macs}
      \vspace{-8pt}
  \end{wrapfigure}
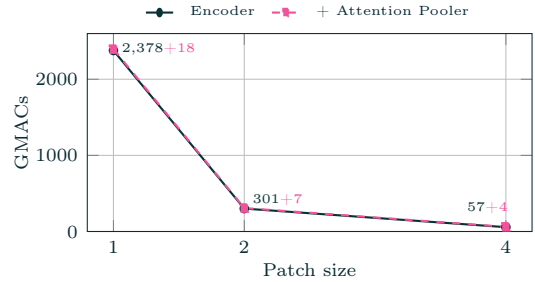

%% file: sec/distillation.tex
\definecolor{oeink}{HTML}{0A3235}
\definecolor{oeteal}{HTML}{105257}
\definecolor{oecream}{HTML}{FAF2E9}
\definecolor{oepink}{HTML}{F0529C}
\colorlet{oepinkfill}{oepink!12}
\begin{figure}[t]
\centering
\resizebox{0.75\linewidth}{!}{%
\begin{tikzpicture}[
    font=\sffamily\footnotesize,
    module/.style={draw=oeteal, line width=0.9pt, fill=oecream, rounded corners=3pt,
                   minimum height=1.0cm, minimum width=2.3cm, align=center, text=oeink},
    loss/.style={module, minimum height=0.5cm, minimum width=1.9cm},
    note/.style={align=center, font=\sffamily\scriptsize, text=oeink},
    arr/.style={-{Triangle[length=2.0mm, width=2.2mm]}, line width=0.9pt, draw=oeteal},
    qcell/.style={draw=oepink, fill=oepinkfill, line width=0.6pt},
    hicell/.style={draw=oepink, fill=oepink, line width=0.6pt},
    panel/.style={draw=oeteal, line width=0.9pt, fill=oecream!60, rounded corners=4pt},
    brace/.style={decorate, decoration={brace, amplitude=5pt, aspect=#1},
                  draw=oeteal, line width=0.8pt},
    brace/.default=0.5,
    sg/.pic={\draw[oepink, line width=1.0pt] (-0.09,-0.13) -- (0.01,0.13) (0.03,-0.13) -- (0.13,0.13);},
  ]
  \foreach \k in {0,...,11}
    \draw[qcell] (0, -1.2 + \k*0.2) rectangle ++(0.35, 0.2);
  \node[note, anchor=north] at (0.175, -1.3) {Location token\\(d768)};

  \node[module] (proj) at (2.45, 0) {Linear +\\LayerNorm\\(768 $\rightarrow$ 128)};
  \draw[arr] (0.5, 0) -- (proj.west);
  \pic at (0.75, 0) {sg};

  \draw[hicell] (4.1, 0) rectangle ++(0.35, 1.2);
  \draw[qcell] (4.1, -1.2) rectangle ++(0.35, 1.2);
  \draw[arr] (proj.east) -- (4.0, 0);
  \node[note, anchor=north] at (4.275, -1.3) {Embedding\\(d128)};

  \draw[brace=0.77] (4.6, 1.2) -- (4.6, -1.2);
  \draw[brace] (4.95, 1.2) -- (4.95, 0.05);

  \foreach \d/\y in {64/0.65, 128/-0.65} {
    \draw[panel] (5.4, \y - 0.6) rectangle (11.15, \y + 0.6);
    \node[text=oeink, font=\sffamily\small] at (5.85, \y) {d\d};
    \node[loss] (mlp\d) at (7.35, \y + 0.27) {MLP};
    \node[loss] (gram\d) at (7.35, \y - 0.27) {Gram loss};
    \node[loss] (cos\d) at (9.9, \y + 0.27) {Cosine loss};
    \draw[arr] (mlp\d.east) -- (cos\d.west);
  }
\end{tikzpicture}%
}
\caption{We use a linear layer to distill the 768-dimensional location tokens to 128-dimensional embeddings. We apply the distillation losses to the 128-dimensional embeddings and to their first 64 dimensions.}
\label{fig:distillation}
\end{figure}
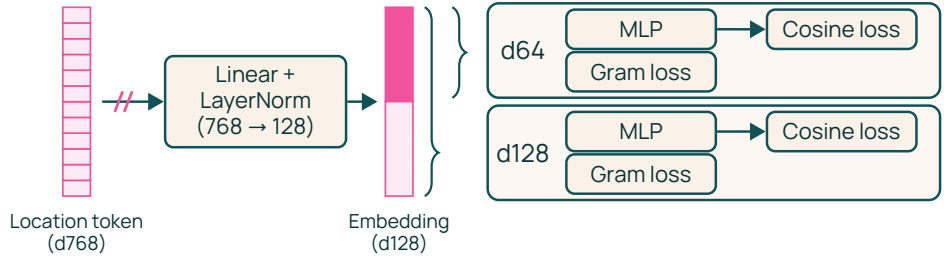

%% file: sec/4_experiments.tex
\section{Experiments \& Results}
\label{sec:experiments}

\subsection{Embedding evaluations}\label{sec:embedding_results}

\input{sec/classifiers_full}

We compute our embeddings using all three input modalities (Sentinel-1, Sentinel-2 and Landsat). We use 12 input timesteps (to obtain annualized embeddings), a $16 \times 16$ input window, an encoder patch size of 2 and an output patch size 1 (yielding embeddings at a 10m $\times$ 10m resolution). In this setting, \textbf{the Attention Pooler is essentially free in terms of compute cost}, costing 7GMACs per window (a 2\% increase from 301GMACs per window without the Attention Pooler to 308GMACs per window with the Attention Pooler). 

We compare our 128-dimensional embeddings with TESSERA-v2 Large \cite{feng2026tessera} and Alpha Earth Foundations \cite{brown2025alphaearthfoundationsembeddingfield} (AEF). When running these evaluations, we aim to reflect the deployed results as closely as possible. For AEF, we retrieved the embeddings using Google Earth Engine \cite{gorelick2017google}. For Tessera v2, we followed the inference instructions in the official github repository\footnote{\url{https://github.com/ucam-eo/tessera/tree/master/tessera_infer_v2}}. For OlmoEarth, we compute embeddings using Sentinel-1, Sentinel-2 and Landsat as inputs with the same processing as pretraining (monthly least-cloudy imagery). \textbf{In all cases, we only consider annualized embeddings} since these reflect the most accessible products for all three models. Each embedding is aligned to its labels' calendar year. \textbf{In all cases, we quantize the embedding} using the deployed quantization method.

\input{sec/pastis_planteur_embeddings}

We evaluate our embeddings following the protocol from AEF. Specifically, for each of the AEF evals (Africa crop mask, Canada Crops \{coarse, fine\}, Descals Oil Palm, Ethiopia Crops, GLanCE, LCMAP land use, US trees) we sampled a balanced training set. For each classifier we evaluate, we train a classifier on this training set and evaluate on all non-sampled points. We do multiple draws per dataset, and report mean accuracy.

We additionally evaluate the embeddings on two crop type classification datasets: PASTIS \cite{garnot2021panoptic} and PLANTEUR \cite{wolters2026planteur}. Like PASTIS, PLANTEUR is based on the Registre parcellaire graphique \cite{etalab2017rpg} but uses labels from Overseas France (Martinique, Réunion, French Guiana, Guadeloupe, Mayotte). PLANTEUR's geographic diversity allows us to asses test performance on one island when classifiers are trained on the other islands (``leave-one-island-out'', LOIO). This allows us to assess the geographic generalization of the embeddings. Embeddings from these larger datasets don't fit in memory, so unlike the AEF evaluations we use a linear probe to assess embedding performance (with the same learning rate grid as Section 3.3 of \cite{olmoearth}).

\input{sec/pastis_vs_gmacs} 

AEF and Tessera v2 are not the same size as OlmoEarth (128 dimensions, 128 bytes at int8). AEF is smaller (64 dimensions, 64 bytes at int8) while Tessera v2 is marginally bigger (128 dimensions but with a magnitude float as well, yielding 132 bytes at int8). 

These evaluated embeddings are not bit-identical to the released embeddings. Unlike the released embeddings, we computed these embeddings by tiling the ViT - the released embeddings use Searchlight for artifact free inference (Section \ref{sec:searchlight}). We measured the difference in performance on PASTIS with and without Searchlight inference, and observed a negligible difference (PASTIS val mIoU was 60.4 with Searchlight compared to 59.8 without). 

\subsubsection{Comparison with AEF and Tessera v2}

Across our embedding evaluations, the OlmoEarth v1.3 embeddings are the best performing. We follow AEF in using $k$NN and a ridge regression with $\lambda = 0$ to evaluate the embeddings; this minimizes the classifiers' hyperparameters, ensuring methods aren't penalized for sub-optimal hyperparameters (Section S4.1 from \cite{brown2025alphaearthfoundationsembeddingfield}). Under this protocol, our d128 embeddings outperform AEF and TESSERA v2 on 7/8 AEF datasets. Our d64 embeddings outperform both on 5 to 7 of 8. The one exception is Ethiopia, where TESSERA v2 consistently outpeforms OlmoEarth v1.3. On crop-type-segmentation, our d128 embeddings slightly outperform TESSERA v2 on PASTIS (58.6 vs. 58.3 mIoU) and are best on PLANTEUR, while our d64 embeddings outperform AEF on PASTIS and both AEF and TESSERA v2 on PLANTEUR.

When deploying models practitioners are likely to experiment with a range of different classifiers. To reflect this, we run the AEF evaluations with a range of commonly used classifiers, using library defaults (Table \ref{tab:aef_classifiers_full}). Different classifiers do not significantly change the overall ranks of the models: our d128 embeddings remain best on 7/8 datasets under every linear classifier and 6/8 under random forests and the MLP. For OlmoEarth and TESSERA, adding some regularization to linear models improves performance - the Ridge ($\lambda = 1$) and Logistic Regression are the best performing classifiers. A small MLP (2 layers, with hidden width 256) also does well; tree-based approaches and $k$NN do worse. AEF does not benefit from regularization but does benefit from more complex classifiers like the MLP and random forest (Table \ref{tab:aef_classifiers_full}). 

\subsubsection{Embedding quality as a function of ViT-patch size} \label{sec:embedding_ps}

Our goal is to produce 10m embeddings. The Attention Pooler allows us to do this with variable patch sizes for the encoder, controlling the computational cost of producing these embeddings. Figure \ref{fig:pastis_vs_gmacs} shows the cost-performance tradeoff of running OlmoEarth v1.3's encoder at patch size 1, 2 and 4 on the PASTIS val set. Running the encoder at patch size 2 dramatically reduces compute cost compared to patch size 1 (a \textbf{7.8 $\times$} decrease, from 2,396 GMACs to 308GMACs) with only a small decrease in performance (from 60.2 mIoU to 59.8 mIoU). This is the setting we use for evaluations (as discussed in Section \ref{sec:embedding_results}) and for deployment.

\subsection{Non-embedding evaluations}\label{sec:nonemb}

\input{sec/ranks_vs_macs_v2}

We also evaluate our model using the same evals as in OlmoEarth v1 and v1.2; this allows us to assess the overall strength of the backbone and Attention Pooler. When running these evaluations, we consider two methods of running OlmoEarth v1.3; with and without the Attention Pooler. When evaluating without the Attention Pooler, we average the Vision Transformer output tokens (following OlmoEarth v1.2). 

The cost of the Attention Pooler is highly dependant on the evaluation being run (and -- most importantly -- on the number of time and modality tokens being compressed). For PASTIS (12 timesteps are compressed into a single token), the Attention Pooler only adds 2\% to the compute cost. For m-eurosat or m-bigearthnet (the worst case -- single timestep, single modality tasks where no compression happens at all), the Attention Pooler adds 16\% to the compute cost. Averaged across all non-embedding eval tasks, the Attention Pooler adds 9\% to the compute cost compared to an unmodified Vision Transformer backbone. 

For these tasks, we use a patch size of 4 for both the encoder and Attention Pooler (unlike the embeddings, which are run at an encoder patch size of 2). 

\subsubsection{kNN and Linear Probing Results} \label{sec:knnlp}

\input{sec/knn_lp_subset}

When evaluated on the $k$NN/LP evaluations, OlmoEarth v1.3 performs slightly better than v1 and v1.2. When ranking across the 13 models which have results on all 24 tasks (AnySat, Clay, Panopticon and all OlmoEarth models), v1.3 Base -- without the Attention Pooler -- has the highest rank, 4.1 compared to 4.3 for OlmoEarth v1.2 and 4.2 for OlmoEarth v1 (Table \ref{tab:knnlp_subset}). The same is true if we rank all models on the 13 Sentinel-2 tasks which all models support (Figure \ref{fig:result_summary}).

Adding the Attention Pooler hurts OlmoEarth v1.3 on these evaluations (by 2.1 points on average). The Attention Pooler does best when there are tokens to aggregate, and spatial context to leverage in the latent-self-attention layer - it does well on PASTIS-S2, which meets both of these requirements. We hypothesize this explains the Attention Pooler's regression overall, since many tasks are missing one of these two components. For example, we observe big drops on single-timestep single-modality tasks with nothing to aggregate (most significantly on MADOS) or tasks with only a single spatial token (CropHarvest, Nandi). 

As discussed above (Section \ref{sec:nonemb}) the Attention Pooler adds a non-neglible cost to forward-passes - this is visible in the increased MACs for Base + P in Figure \ref{fig:result_summary}. However, we note again that many of these tasks represent worst-case scenarios for the Attention Pooler's cost, and the Attention Pooler's relative cost for our embeddings evaluations (Section \ref{sec:embedding_results}) is much smaller.

\subsubsection{Finetuning Results} \label{sec:finetuning_results}

\input{sec/finetune_v2}

OlmoEarth v1.3 is a strong finetuning model (Table \ref{tab:finetune}). If we rank all models over all tasks (omitting missing task-model combinations), OlmoEarth v1.3 (with no Attention Pooling) has a mean rank of 4.34. This is behind OlmoEarth v1.2 Base (4.02) and ahead of OlmoEarth v1 Base (4.74). All three of these models are ahead of the non-OlmoEarth models (next best is Galileo-Base at 9.61) and of non-Base OlmoEarth models (next best is OlmoEarth v1.2 Small at 4.80). Like in the $k$NN / LP case, Attention Pooling penalizes the model, lowering its mean rank to 5.88. Attention Pooling tends to hurt most on single-image, single-modality tasks with no tokens to pool (m-so2sat, the partner vessel detection, length and type tasks). 

v1.3 performs well on both research tasks and the partner tasks, ranking second on each (mean ranks 4.50 and 4.23). In contrast, v1.2 is best on the research benchmarks (2.60) but fourth on the partner tasks (4.97) while the opposite is true for v1 (first on partner tasks at 3.03, sixth on research benchmarks at 7.30). 

\subsection{Summary}

OlmoEarth v1.3 produces state-of-the-art embeddings at 128 and 64 dimensions while remaining a strong encoder backbone for frozen and finetuned evaluations. 

In spite of the additional parameters introduced by the Attention Pooler (as well as from the supervision heads and the distillation losses' MLPs), OlmoEarth v1.3 is considerably cheaper to train than OlmoEarth v1 or v1.2 while matching the previous models' performances.

The Attention Pooler lets us produce 10m embeddings at different encoder patch sizes, balancing performance and inference cost. Encoder patch size 2 keeps nearly all of patch size 1's performance while reducing inference costs by $7.8 \times$. This is the configuration we use to compute global embeddings, described below.

%% file: sec/classifiers_full.tex
\sethlcolor{colorone!25}
\begin{table*}[htpb]
    \centering
    \resizebox{\textwidth}{!}{%
    \begin{tabular}{l|cccc|cccc|cccc|cccc|cccc|cccc|cccc|cccc}
 & \multicolumn{4}{c|}{Africa crop mask} & \multicolumn{4}{c|}{Canada crops (coarse)} & \multicolumn{4}{c|}{Canada crops (fine)} & \multicolumn{4}{c|}{Descals oil palm} & \multicolumn{4}{c|}{Ethiopia crops} & \multicolumn{4}{c|}{GLanCE} & \multicolumn{4}{c|}{LCMAP land use} & \multicolumn{4}{c}{US trees} \\
 & AEF & TES. & \multicolumn{2}{c|}{v1.3} & AEF & TES. & \multicolumn{2}{c|}{v1.3} & AEF & TES. & \multicolumn{2}{c|}{v1.3} & AEF & TES. & \multicolumn{2}{c|}{v1.3} & AEF & TES. & \multicolumn{2}{c|}{v1.3} & AEF & TES. & \multicolumn{2}{c|}{v1.3} & AEF & TES. & \multicolumn{2}{c|}{v1.3} & AEF & TES. & \multicolumn{2}{c}{v1.3} \\
Readout & 64 & 128 & 128 & 64 & 64 & 128 & 128 & 64 & 64 & 128 & 128 & 64 & 64 & 128 & 128 & 64 & 64 & 128 & 128 & 64 & 64 & 128 & 128 & 64 & 64 & 128 & 128 & 64 & 64 & 128 & 128 & 64 \\
\hline
Ridge, $\lambda{=}0$ & 88.8 & 88.1 & \cellcolor{colorone!25} 91.9 & \cellcolor{colorone!25} \textbf{92.6} & 70.3 & 70.7 & \cellcolor{colorone!25} \textbf{73.6} & \cellcolor{colorone!25} 73.3 & 60.6 & 64.5 & \cellcolor{colorone!25} \textbf{66.2} & 63.5 & 77.4 & 69.8 & \cellcolor{colorone!25} \textbf{79.1} & \cellcolor{colorone!25} 79.0 & 68.5 & 68.1 & 63.6 & \cellcolor{colorone!25} \textbf{69.1} & 67.7 & 68.7 & \cellcolor{colorone!25} \textbf{70.9} & \cellcolor{colorone!25} 69.5 & 85.8 & 85.2 & \cellcolor{colorone!25} \textbf{88.7} & \cellcolor{colorone!25} 88.1 & 21.1 & 20.9 & \cellcolor{colorone!25} \textbf{23.0} & \cellcolor{colorone!25} 21.2 \\
kNN-5 & 87.6 & 88.3 & \cellcolor{colorone!25} \textbf{91.7} & \cellcolor{colorone!25} 91.6 & 71.2 & 71.0 & \cellcolor{colorone!25} \textbf{72.4} & \cellcolor{colorone!25} 71.5 & 61.3 & 62.6 & \cellcolor{colorone!25} \textbf{64.7} & \cellcolor{colorone!25} 64.0 & 71.2 & 69.0 & \cellcolor{colorone!25} \textbf{72.0} & \cellcolor{colorone!25} 71.6 & 66.4 & \textbf{77.4} & 70.8 & 69.2 & 68.0 & 68.0 & \cellcolor{colorone!25} \textbf{70.2} & \cellcolor{colorone!25} 70.0 & 83.4 & 83.7 & \cellcolor{colorone!25} 86.3 & \cellcolor{colorone!25} \textbf{86.3} & 18.3 & 16.1 & \cellcolor{colorone!25} \textbf{18.6} & 18.3 \\
kNN-20 & 86.9 & 87.1 & \cellcolor{colorone!25} \textbf{91.8} & \cellcolor{colorone!25} 91.6 & 70.9 & 70.6 & \cellcolor{colorone!25} \textbf{71.7} & \cellcolor{colorone!25} 70.9 & 61.9 & 63.3 & \cellcolor{colorone!25} \textbf{64.9} & \cellcolor{colorone!25} 63.9 & 72.2 & 70.3 & \cellcolor{colorone!25} \textbf{73.2} & \cellcolor{colorone!25} 72.8 & 64.3 & \textbf{74.2} & 67.0 & 66.2 & 68.8 & 68.8 & \cellcolor{colorone!25} \textbf{71.0} & \cellcolor{colorone!25} 70.8 & 84.2 & 84.2 & \cellcolor{colorone!25} \textbf{87.1} & \cellcolor{colorone!25} 87.0 & 21.9 & 19.5 & \cellcolor{colorone!25} \textbf{22.0} & 21.6 \\
\hline
Ridge, $\lambda{=}1$ & 89.1 & 89.7 & \cellcolor{colorone!25} \textbf{93.2} & \cellcolor{colorone!25} 93.0 & 70.1 & 71.5 & \cellcolor{colorone!25} \textbf{74.5} & \cellcolor{colorone!25} 73.3 & 60.1 & 64.5 & \cellcolor{colorone!25} \textbf{66.0} & 63.1 & 76.4 & 71.5 & \cellcolor{colorone!25} \textbf{80.6} & \cellcolor{colorone!25} 78.4 & 66.2 & \textbf{79.6} & 73.8 & 70.3 & 67.7 & 68.7 & \cellcolor{colorone!25} \textbf{70.9} & \cellcolor{colorone!25} 69.5 & 85.7 & 85.2 & \cellcolor{colorone!25} \textbf{88.7} & \cellcolor{colorone!25} 88.1 & 21.1 & 20.9 & \cellcolor{colorone!25} \textbf{22.9} & \cellcolor{colorone!25} 21.3 \\
Ridge, $\lambda{=}10$ & 86.8 & 90.1 & \cellcolor{colorone!25} \textbf{93.3} & \cellcolor{colorone!25} 92.9 & 68.1 & 71.5 & \cellcolor{colorone!25} \textbf{74.0} & \cellcolor{colorone!25} 72.3 & 57.6 & 64.1 & \cellcolor{colorone!25} \textbf{65.1} & 62.3 & 72.9 & 72.4 & \cellcolor{colorone!25} \textbf{79.8} & \cellcolor{colorone!25} 77.0 & 60.2 & \textbf{80.1} & 67.1 & 64.9 & 67.4 & 68.6 & \cellcolor{colorone!25} \textbf{70.8} & \cellcolor{colorone!25} 69.4 & 85.2 & 85.2 & \cellcolor{colorone!25} \textbf{88.7} & \cellcolor{colorone!25} 88.1 & 20.8 & 20.8 & \cellcolor{colorone!25} \textbf{22.8} & \cellcolor{colorone!25} 21.2 \\
LogReg & 87.9 & 89.8 & \cellcolor{colorone!25} \textbf{93.0} & \cellcolor{colorone!25} 92.8 & 69.3 & 70.4 & \cellcolor{colorone!25} \textbf{74.4} & \cellcolor{colorone!25} 73.4 & 62.0 & 64.5 & \cellcolor{colorone!25} \textbf{66.8} & \cellcolor{colorone!25} 65.7 & 75.6 & 71.6 & \cellcolor{colorone!25} \textbf{81.3} & \cellcolor{colorone!25} 79.1 & 63.3 & \textbf{79.7} & 71.0 & 67.9 & 69.1 & 68.7 & \cellcolor{colorone!25} \textbf{71.0} & \cellcolor{colorone!25} 70.9 & 86.2 & 84.6 & \cellcolor{colorone!25} \textbf{88.4} & \cellcolor{colorone!25} 88.2 & 22.3 & 22.0 & \cellcolor{colorone!25} \textbf{23.9} & \cellcolor{colorone!25} 22.9 \\
RF & 88.5 & 89.6 & \cellcolor{colorone!25} \textbf{92.5} & \cellcolor{colorone!25} 92.0 & 71.0 & 71.5 & \cellcolor{colorone!25} \textbf{72.7} & \cellcolor{colorone!25} 71.9 & 62.7 & 64.7 & \cellcolor{colorone!25} \textbf{65.4} & \cellcolor{colorone!25} 64.7 & 74.3 & 71.7 & \cellcolor{colorone!25} \textbf{77.9} & \cellcolor{colorone!25} 77.0 & 68.8 & \textbf{77.3} & 71.3 & 70.1 & 69.2 & 69.5 & \cellcolor{colorone!25} \textbf{71.3} & \cellcolor{colorone!25} 71.0 & 84.4 & 84.1 & \cellcolor{colorone!25} \textbf{87.0} & \cellcolor{colorone!25} 86.8 & \textbf{22.4} & 20.3 & 21.9 & 21.5 \\
MLP & 89.6 & 90.1 & \cellcolor{colorone!25} \textbf{92.9} & \cellcolor{colorone!25} 92.5 & 71.2 & 71.7 & \cellcolor{colorone!25} \textbf{73.7} & \cellcolor{colorone!25} 73.2 & 62.2 & 63.1 & \cellcolor{colorone!25} \textbf{65.7} & \cellcolor{colorone!25} 64.5 & 76.6 & 72.7 & \cellcolor{colorone!25} \textbf{80.5} & \cellcolor{colorone!25} 78.8 & 71.9 & \textbf{80.4} & 75.9 & 74.0 & 68.3 & 67.0 & \cellcolor{colorone!25} \textbf{69.9} & \cellcolor{colorone!25} 69.2 & 85.4 & 84.4 & \cellcolor{colorone!25} \textbf{87.8} & \cellcolor{colorone!25} 87.6 & \textbf{17.6} & 15.3 & 17.2 & 17.5 \\
    \end{tabular}}
    \caption{Balanced accuracy on the eight AEF balanced-trial datasets for every readout. Sub-columns are AEF (64 dims), TESSERA v2 (``TES.'', 128 dims) and {\model} v1.3 at 128 dims and its first 64 dims. The first block is the AEF protocol (ridge with $\lambda{=}0$, kNN with $k{=}5, 20$); the second block adds commonly used classifiers at library defaults. Every classifier is fit on the same class-balanced draws at AEF's Table-1 sizes and scored on the remainder; cells are means over repeated draws. We \hl{highlight} {\model} cells that beat the better of the two baselines.}
    \label{tab:aef_classifiers_full}
\end{table*}

%% file: sec/pastis_planteur_embeddings.tex
\begin{wraptable}{r}{0.5\linewidth}
    \vspace{-8pt}
    \centering
    \resizebox{\linewidth}{!}{
    \begin{tabular}{ll|c|c|c}
 &  & PASTIS & \multicolumn{2}{c}{PLANTEUR}\\
Model & Dim. & Random & Random & LOIO \\
\hline
AEF & 64 & 39.1 & 27.7 & 15.8 \\
TESSERA v2 & 128 & 58.3 & 34.2 & 19.1 \\
\hline
{\model} v1.3 distilled & 128 & \cellcolor{colorone!25} \textbf{58.6} & \cellcolor{colorone!25} \textbf{36.7} & \cellcolor{colorone!25} \textbf{24.9} \\
{\model} v1.3 distilled & 64 & 53.4 & \cellcolor{colorone!25} 35.2 & 21.5 \\
    \end{tabular}}
    \vspace{-2pt}
    \caption{Crop-type segmentation results, measured via mIoU, on the PASTIS and PLANTEUR datasets. We also measure PLANTEUR's ``leave-one-island-out'' (LOIO) mIoU performance, which evaluates the embeddings' geographic generalization capabilities.}
    \label{tab:pastis_planteur}
    \vspace{-8pt}
\end{wraptable}

%% file: sec/pastis_vs_gmacs.tex
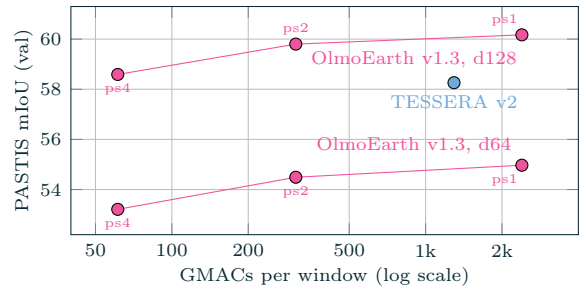
\begin{wrapfigure}{r}{0.5\linewidth}
    \vspace{-8pt}
    \centering
    \begin{tikzpicture}
        \tikzset{modelmarker/.style={circle, inner sep=1.6pt, draw=black}}
        \begin{axis}[
            width=\linewidth,
            height=4.6cm,
            clip=false,
            xmode=log,
            log basis x=10,
            xmin=40, xmax=4000,
            ymin=52.2, ymax=61.3,
            xtick={50, 100, 200, 500, 1000, 2000},
            xticklabels={50, 100, 200, 500, 1k, 2k},
            ytick={54, 56, 58, 60},
            xlabel={GMACs per window (log scale)},
            ylabel={PASTIS mIoU (val)},
            xlabel style={font=\scriptsize, yshift=3pt},
            ylabel style={font=\scriptsize, yshift=-4pt},
            tick label style={font=\scriptsize},
            grid=major,
        ]
        \coordinate (ps4)  at (axis cs:61.2,   58.59);
        \coordinate (ps2)  at (axis cs:307.9,  59.80);
        \coordinate (ps1)  at (axis cs:2395.5, 60.17);
        \coordinate (d64ps4) at (axis cs:61.2,   53.21);
        \coordinate (d64ps2) at (axis cs:307.9,  54.49);
        \coordinate (d64ps1) at (axis cs:2395.5, 54.97);
        \coordinate (tes)  at (axis cs:1293.9, 58.26);

        \draw[colorone,-] (ps4) -- (ps2) -- (ps1);
        \draw[colorone,-] (d64ps4) -- (d64ps2) -- (d64ps1);

        \node[modelmarker, fill=colorone] at (ps4) {};
        \node[modelmarker, fill=colorone] at (ps2) {};
        \node[modelmarker, fill=colorone] at (ps1) {};
        \node[anchor=north, text=colorone] at (ps4) {\tiny ps4};
        \node[anchor=south, text=colorone] at (ps2) {\tiny ps2};
        \node[anchor=south east, text=colorone, xshift=2pt] at (ps1) {\tiny ps1};
        \node[anchor=north, text=colorone] at (axis cs:900, 59.88) {\scriptsize {\model} v1.3, d128};

        \node[modelmarker, fill=colorone] at (d64ps4) {};
        \node[modelmarker, fill=colorone] at (d64ps2) {};
        \node[modelmarker, fill=colorone] at (d64ps1) {};
        \node[anchor=north, text=colorone] at (d64ps4) {\tiny ps4};
        \node[anchor=north, text=colorone] at (d64ps2) {\tiny ps2};
        \node[anchor=north east, text=colorone, xshift=2pt] at (d64ps1) {\tiny ps1};
        \node[anchor=south, text=colorone] at (axis cs:900, 55.05) {\scriptsize {\model} v1.3, d64};

        \node[modelmarker, fill=colorthree] at (tes) {};
        \node[anchor=north, text=colorthree, yshift=-1pt] at (tes) {\scriptsize TESSERA v2};
        \end{axis}
    \end{tikzpicture}
    \vspace{-6pt}
    \caption{PASTIS validation mIoU against inference cost per 16$\times$16$\times$12 S1+S2+L8 window, for the 128-dimensional embeddings and their first 64 dimensions (d64). In all cases, the model's output patch size is 1 (10m). AEF scores 39.1.}
    \label{fig:pastis_vs_gmacs}
    \vspace{-8pt}
\end{wrapfigure}

%% file: sec/ranks_vs_macs_v2.tex
  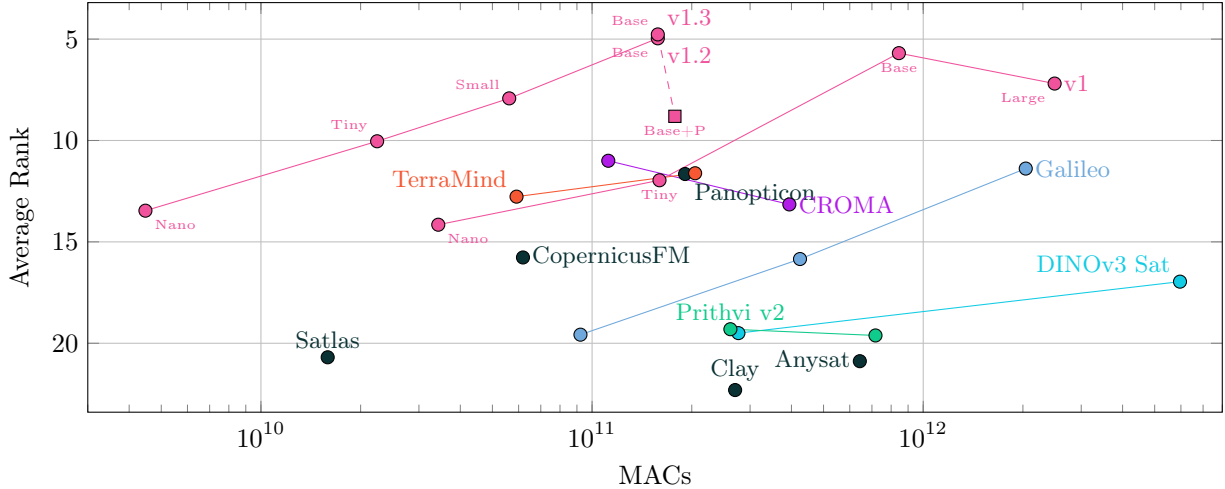
\begin{figure}[t]
      \centering
      \begin{tikzpicture}
          \tikzset{modelmarker/.style={circle, inner sep=1.73329pt, draw=black}}

          \begin{axis}[
              xmode=log,
              width=\columnwidth,
              height=7cm,
              xmin=3e9, xmax=8e12,
              ymin=3.2, ymax=23.4,
              xtick={1e9, 1e10, 1e11, 1e12, 1e13},
              xlabel={MACs},
              ylabel={Average Rank},
              grid=major,
              y dir=reverse
          ]

            \coordinate (croma-1)      at (axis cs:111924196549, 11.00000000);
            \coordinate (croma-2)      at (axis cs:394294330762, 13.15384615);
            \coordinate (dinov3-1)     at (axis cs:276638507638, 19.50000000);
            \coordinate (dinov3-2)     at (axis cs:5963011946023, 16.96153846);
            \coordinate (galileo-1)    at (axis cs:92241634658, 19.57692308);
            \coordinate (galileo-2)    at (axis cs:424512375434, 15.84615385);
            \coordinate (galileo-3)    at (axis cs:2039829824197, 11.38461538);
            \coordinate (prithvi-1)    at (axis cs:261633664866, 19.30769231);
            \coordinate (prithvi-2)    at (axis cs:717538667914, 19.61538462);
            \coordinate (terramind-1)  at (axis cs:59136953738, 12.76923077);
            \coordinate (terramind-2)  at (axis cs:204739722004, 11.61538462);
            \coordinate (v1-1)         at (axis cs:34316909647, 14.15384615);
            \coordinate (v1-2)         at (axis cs:159805931993, 11.96153846);
            \coordinate (v1-3)         at (axis cs:844292363815, 5.69230769);
            \coordinate (v1-4)         at (axis cs:2493645662050, 7.19230769);
            \coordinate (v12-1)        at (axis cs:4479262247, 13.46153846);
            \coordinate (v12-2)        at (axis cs:22432871188, 10.03846154);
            \coordinate (v12-3)        at (axis cs:56205749642, 7.92307692);
            \coordinate (v12-4)        at (axis cs:157929759980, 4.96153846);
            \coordinate (v13-1)        at (axis cs:157929759980, 4.76923077);
            \coordinate (v13-2)        at (axis cs:177865820396, 8.80769231);
            \coordinate (anysat)       at (axis cs:643495127188, 20.88461538);
            \coordinate (clay)         at (axis cs:270365348805, 22.30769231);
            \coordinate (copernicusfm) at (axis cs:61847048921, 15.76923077);
            \coordinate (panopticon)   at (axis cs:190944609674, 11.65384615);
            \coordinate (satlas)       at (axis cs:15906683353, 20.69230769);

          \draw[colortwo,-]   (croma-1) -- (croma-2);
          \draw[colorfive,-]  (dinov3-1) -- (dinov3-2);
          \draw[colorthree,-] (galileo-1) -- (galileo-2) -- (galileo-3);
          \draw[colorfour,-]  (prithvi-1) -- (prithvi-2);
          \draw[colorsix,-]   (terramind-1) -- (terramind-2);
          \draw[colorone,-]   (v1-1) -- (v1-2) -- (v1-3) -- (v1-4);
          \draw[colorone,-]   (v12-1) -- (v12-2) -- (v12-3) -- (v12-4);
          \draw[colorone,dashed] (v13-1) -- (v13-2);

          \node[modelmarker, fill=colordefault] at (anysat) {};
          \node[anchor=east,  text=colordefault] at (anysat)       {\small Anysat };
          \node[modelmarker, fill=colordefault] at (clay) {};
          \node[anchor=south, text=colordefault] at (clay)         {\small Clay };
          \node[modelmarker, fill=colordefault] at (copernicusfm) {};
          \node[anchor=west, text=colordefault] at (copernicusfm) {\small CopernicusFM };
          \node[modelmarker, fill=colordefault] at (panopticon) {};
          \node[anchor=north west, text=colordefault] at (panopticon)   {\small Panopticon };
          \node[modelmarker, fill=colordefault] at (satlas) {};
          \node[anchor=south, text=colordefault] at (satlas)       {\small Satlas };

          \node[modelmarker, fill=colortwo] at (croma-1) {};
          \node[modelmarker, fill=colortwo] at (croma-2) {};
          \node[anchor=west, text=colortwo] at (croma-2) {\small CROMA };

          \node[modelmarker, fill=colorfive] at (dinov3-1) {};
          \node[modelmarker, fill=colorfive] at (dinov3-2) {};
          \node[anchor=south east, text=colorfive] at (dinov3-2) {\small DINOv3 Sat };

          \node[modelmarker, fill=colorthree] at (galileo-1) {};
          \node[modelmarker, fill=colorthree] at (galileo-2) {};
          \node[modelmarker, fill=colorthree] at (galileo-3) {};
          \node[anchor=west, text=colorthree] at (galileo-3) {\small Galileo };

          \node[modelmarker, fill=colorfour] at (prithvi-1) {};
          \node[modelmarker, fill=colorfour] at (prithvi-2) {};
          \node[anchor=south, text=colorfour] at (prithvi-1) {\small Prithvi v2 };

          \node[modelmarker, fill=colorsix] at (terramind-1) {};
          \node[modelmarker, fill=colorsix] at (terramind-2) {};
          \node[anchor=south east, text=colorsix] at (terramind-1) {\small TerraMind };

          \node[modelmarker, fill=colorone] at (v1-1) {};
          \node[modelmarker, fill=colorone] at (v1-2) {};
          \node[modelmarker, fill=colorone] at (v1-3) {};
          \node[modelmarker, fill=colorone] at (v1-4) {};
          \node[anchor=west, text=colorone] at (v1-4) {\small v1};

          \node[modelmarker, fill=colorone] at (v12-1) {};
          \node[modelmarker, fill=colorone] at (v12-2) {};
          \node[modelmarker, fill=colorone] at (v12-3) {};
          \node[modelmarker, fill=colorone] at (v12-4) {};
          \node[anchor=north west, text=colorone] at (v12-4) {\small v1.2};

          \node[modelmarker, fill=colorone] at (v13-1) {};
          \node[rectangle, draw=black, fill=colorone, inner sep=2.3pt] at (v13-2) {};
          \node[anchor=south west, text=colorone] at (v13-1) {\small v1.3};

          \node[anchor=north west, text=colorone] at (v12-1) {\tiny Nano };
          \node[anchor=south east, text=colorone] at (v12-2) {\tiny Tiny };
          \node[anchor=south east, text=colorone] at (v12-3) {\tiny Small };
          \node[anchor=north east, text=colorone] at (v12-4) {\tiny Base };

          \node[anchor=north west, text=colorone] at (v1-1) {\tiny Nano };
          \node[anchor=north, text=colorone] at (v1-2) {\tiny Tiny };
          \node[anchor=north, text=colorone] at (v1-3) {\tiny Base };
          \node[anchor=north east, text=colorone] at (v1-4) {\tiny Large};

          \node[anchor=south east, text=colorone] at (v13-1) {\tiny Base };
          \node[anchor=north, text=colorone] at (v13-2) {\tiny Base+P};

          \end{axis}
      \end{tikzpicture}
      \caption{Performance vs. computational efficiency, averaged across 13 $k$NN/LP tasks (Sentinel-2 tasks which all models can support). The chart shows average Multiply-Accumulate operations (MACs) to encode one example across all tasks (input size varies by task) - each line connects model sizes within a family. We measure OlmoEarth v1.3 performance without the Attention Pooler and with (+P, square). Without the Attention Pooler, v1.3 has the best average rank of all models, at the same cost as v1.2 Base; on these 13 tasks, the Attention Pooler adds 13\% to the MACs. Full results in Table \ref{tab:knnlp}.}
      \label{fig:result_summary}
  \end{figure}

%% file: sec/knn_lp_subset.tex
\sethlcolor{colorone!25}
\begin{table*}[htpb]
    \centering
    \resizebox{\textwidth}{!}{%


    \begin{tabular}{ll|cccccccccccccccccc|cccccc}

\hdr{} & \hdr{} & \hdr{m-bigearthnet} & \hdr{m-so2sat} & \hdr{m-brick-kiln} & \hdr{m-forestnet} & \hdr{m-eurosat} & \hdr{BreizhCrops} & \hdr{CropHarvest-PRC} & \hdr{CropHarvest-PRC} & \hdr{CropHarvest-PRC} & \hdr{CropHarvest-Togo} & \hdr{CropHarvest-Togo} & \hdr{CropHarvest-Togo} & \hdr{m-cashewplant} & \hdr{m-SA-crop-type} & \hdr{PASTIS} & \hdr{PASTIS} & \hdr{MADOS} & \hdr{Sen1Floods11} & \hdr{AWF} & \hdr{AWF} & \hdr{AWF} & \hdr{Nandi} & \hdr{Nandi} & \hdr{Nandi} \\
 & Modalities & S2 & S2 & S2 & L8 & S2 & S2 & S1 & S2 & S1,S2 & S1 & S2 & S1,S2 & S2 & S2 & S1 & S2 & S2 & S1 & L8 & S1 & S2 & L8 & S1 & S2 \\
 & Time series & \ccross & \ccross & \ccross & \ccross & \ccross & \ccheck & \ccheck & \ccheck & \ccheck & \ccheck & \ccheck & \ccheck & \ccross & \ccross & \ccheck & \ccheck & \ccross & \ccross & \ccheck & \ccheck & \ccheck & \ccheck & \ccheck & \ccheck \\
 & Method & kNN & kNN & kNN & kNN & kNN & LP & LP & LP & LP & LP & LP & LP & LP & LP & LP & LP & LP & LP & kNN & kNN & kNN & kNN & kNN & kNN \\
Model & Metric & $\mu \text{F1}$ & Acc. & Acc. & Acc. & Acc. & Acc. & Acc. & Acc. & Acc. & Acc. & Acc. & Acc. & mIOU & mIOU & mIOU & mIOU & mIOU & mIOU & Acc. & Acc. & Acc. & Acc. & Acc. & Acc. \\
\hline
Best non-{\model} &  & 64.9 & 60.5 & 94.1 & \textbf{52.3} & 95.0 & 69.7 & 60.9 & \textbf{81.9} & \textbf{79.3} & 79.4 & 81.4 & 81.0 & \textbf{54.1} & 31.7 & 28.0 & 44.7 & \textbf{68.4} & 79.4 & 66.0 & 67.5 & \textbf{79.0} & 61.5 & 26.8 & 68.2 \\
\hline
{\model} v1 & ViT Base & 62.4 & \cellcolor{colorone!25} \textbf{67.7} & 94.3 & 41.9 & \cellcolor{colorone!25} \textbf{95.5} & 70.9 & 56.8 & 73.4 & 75.4 & \cellcolor{colorone!25} \textbf{80.1} & \cellcolor{colorone!25} \textbf{87.3} & \cellcolor{colorone!25} \textbf{82.0} & \cellcolor{colorone!25} 32.3 & 28.9 & 29.7 & 50.6 & \cellcolor{colorone!25} 67.2 & 79.2 & \cellcolor{colorone!25} \textbf{77.0} & \cellcolor{colorone!25} \textbf{68.5} & \cellcolor{colorone!25} 77.5 & \cellcolor{colorone!25} \textbf{67.9} & 26.5 & 74.7 \\
{\model} v1.2 & ViT Base & 65.5 & 67.3 & 94.7 & \cellcolor{colorone!25} 45.4 & 95.0 & \cellcolor{colorone!25} \textbf{73.3} & 58.2 & 76.7 & 77.7 & 74.8 & 81.4 & 81.0 & 29.6 & 32.0 & 31.1 & 54.2 & \cellcolor{colorone!25} 67.2 & 78.9 & 74.5 & 68.0 & 75.0 & 65.7 & 28.1 & 73.9 \\
\hline
{\model} v1.3 & ViT Base + P & \cellcolor{colorone!25} 65.9 & 55.4 & \cellcolor{colorone!25} 95.3 & 37.3 & 94.1 & 70.9 & 57.1 & 76.5 & 76.9 & 74.5 & 77.1 & 79.7 & 28.9 & 27.7 & \cellcolor{colorone!25} 31.4 & \cellcolor{colorone!25} \textbf{57.5} & 51.7 & \cellcolor{colorone!25} \textbf{79.5} & 73.0 & 65.0 & 76.0 & 66.4 & 27.0 & 71.5 \\
{\model} v1.3 & ViT Base & \cellcolor{colorone!25} \textbf{66.1} & 58.2 & \cellcolor{colorone!25} \textbf{96.8} & 41.7 & 93.4 & 73.2 & \cellcolor{colorone!25} \textbf{62.4} & \cellcolor{colorone!25} 77.7 & \cellcolor{colorone!25} 78.3 & 72.9 & 82.0 & 80.1 & 29.1 & \cellcolor{colorone!25} \textbf{32.8} & \cellcolor{colorone!25} \textbf{32.6} & \cellcolor{colorone!25} 54.4 & 67.1 & \cellcolor{colorone!25} 79.3 & 72.0 & 67.0 & 76.0 & 67.4 & \cellcolor{colorone!25} \textbf{28.5} & \cellcolor{colorone!25} \textbf{77.6} \\
\end{tabular}}
    \caption{kNN/Linear probe results on research benchmarks and real-world partner tasks for the {\model} Base models. ``Best non-{\model}'' is, for each task, the best score among all other models we evaluate (full results in Table~\ref{tab:knnlp}). We run v1.3 with the Attention Pooler (``+P'') and without. We \textbf{bold} the best score among the rows shown and \hl{highlight} which ViT Base model wins between v1, v1.2, and v1.3.}
    \label{tab:knnlp_subset}
\end{table*}

%% file: sec/finetune_v2.tex
\sethlcolor{colorone!25}
\begin{table*}[htpb]
    \centering
    \resizebox{\textwidth}{!}{%
    \begin{tabular}{ll|cccccccccc|ccccccccccccccccc}

\hdr{} & \hdr{} & \hdr{m-bigearthnet} & \hdr{m-so2sat} & \hdr{m-brick-kiln} & \hdr{m-forestnet} & \hdr{m-eurosat} & \hdr{m-cashewplant} & \hdr{m-SA-crop-type} & \hdr{PASTIS} & \hdr{MADOS} & \hdr{Sen1Floods11} & \hdr{AWF} & \hdr{AWF} & \hdr{GEA North Africa} & \hdr{Forest Loss Driver} & \hdr{Live Fuel Moisture Content} & \hdr{Live Fuel Moisture Content} & \hdr{Mangrove} & \hdr{Mangrove} & \hdr{Nandi} & \hdr{Nandi} & \hdr{Vessel Detection} & \hdr{Vessel Detection} & \hdr{Vessel Length} & \hdr{Vessel Type} & \hdr{Solar Farm Detection} \\
 & Modalities & S2 & S2 & S2 & L8 & S2 & S2 & S2 & S2 & S2 & S1 & S2 & S2, S1 & S2 & S2 & S2 & S2, S1 & S2 & S2, S1 & S2 & S2, S1 & L8 & S1 & S2 & S2 & S2 \\
 & Time series & \ccross & \ccross & \ccross & \ccross & \ccross & \ccross & \ccross & \ccheck & \ccross & \ccross & \ccheck & \ccheck & \ccheck & \ccheck & \ccheck & \ccheck & \ccheck & \ccheck & \ccheck & \ccheck & \ccross & \ccross & \ccross & \ccross & \ccheck \\
Model & Metric & $\mu \text{F1}$ & Acc. & Acc. & Acc. & Acc. & mIOU & mIOU & mIOU & mIOU & mIOU & Acc. & Acc. & Acc. & Acc. & L1 & L1 & Acc. & Acc. & Acc. & Acc. & F1 & F1 & L1 & Acc. & mIoU \\
\hline
Anysat & ViT Base & 68.4 & 56.7 & 98.7 & 51.6 & 95.9 & 80.4 & 34.2 & 60.9 & 63.3 & 77.4 & 78.0 & 83.0 & 59.3 & 84.6 & 19.1 & 19.4 & 96.9 & 97.1 & 78.4 & 76.7 & - & - & 73.4 & 43.5 & 82.6 \\
Clay & ViT Large & 65.7 & 61.3 & 98.2 & 49.2 & 95.1 & 73.9 & 33.4 & 48.9 & 68.9 & 78.5 & 77.0 & - & 53.4 & 93.3 & 24.8 & - & 96.6 & - & 30.9 & - & 70.7 & 79.9 & 16.4 & 68.3 & 82.2 \\
CopernicusFM & ViT Base & 71.3 & 66.8 & 98.3 & - & 98.4 & 78.7 & 33.6 & 54.6 & 66.0 & 78.6 & 79.0 & 79.0 & 58.8 & 90.0 & 25.2 & 24.5 & 97.1 & 97.1 & 68.2 & 55.9 & - & 77.4 & 16.7 & 69.6 & 77.6 \\
CROMA & ViT Base & 69.5 & 59.1 & 98.7 & - & 95.6 & 46.4 & 34.8 & 56.3 & 66.6 & 79.4 & 76.5 & 75.5 & 57.5 & 93.2 & 24.3 & 24.0 & 96.4 & 96.3 & 62.2 & 67.4 & - & - & 19.8 & 64.4 & 79.5 \\
CROMA & ViT Large & 71.8 & 58.9 & 97.8 & - & 97.5 & 47.8 & 36.0 & 58.1 & 68.8 & 79.4 & 79.5 & - & 56.6 & 92.3 & 24.6 & 24.1 & 96.6 & 96.2 & 76.4 & - & - & - & - & - & 81.7 \\
DINOv3 Sat & ViT Large & 69.9 & 63.3 & 98.9 & \textbf{59.2} & 96.7 & 80.6 & 34.5 & 42.8 & 64.7 & - & 34.5 & - & 43.0 & 80.4 & 90.2 & - & 65.8 & - & 35.8 & - & - & - & 30.3 & 54.8 & 70.7 \\
Galileo & ViT Base & 69.2 & 64.7 & 98.3 & - & 97.8 & 78.8 & 35.7 & 61.2 & 71.9 & 79.7 & 81.0 & 81.5 & 62.9 & 95.1 & 20.1 & 18.7 & 97.3 & 97.5 & \textbf{81.9} & 81.9 & - & 78.7 & 16.4 & 73.0 & 83.1 \\
Panopticon & ViT Base & 69.3 & 65.4 & 99.0 & 56.0 & 98.2 & 79.7 & 33.4 & 54.4 & 72.8 & 79.1 & 75.5 & 78.5 & 54.3 & 96.4 & 24.5 & 23.7 & 97.1 & 97.4 & 65.2 & 69.5 & 74.9 & 76.7 & 17.7 & 69.4 & 81.8 \\
Prithvi v2 & ViT Huge & 70.6 & 64.7 & 98.2 & - & 96.8 & 81.1 & 38.8 & 58.6 & 69.3 & - & 80.0 & - & 60.6 & 92.4 & - & - & 97.2 & - & 77.1 & - & 71.1 & - & 17.4 & 68.2 & 84.1 \\
Satlas & Swin Base & 72.7 & 65.1 & 98.7 & 56.0 & 97.0 & 77.0 & 37.8 & 57.4 & 60.5 & 78.5 & 78.0 & - & 56.1 & 63.3 & 25.0 & 24.6 & 96.6 & - & 47.6 & - & - & - & 16.2 & 71.6 & 83.3 \\
TerraMind & ViT Base & 72.6 & 66.1 & 98.5 & - & 97.6 & 80.9 & 39.2 & 59.9 & 73.2 & 79.5 & 84.0 & 82.0 & 49.8 & 96.4 & 24.3 & 23.8 & \textbf{97.7} & 96.8 & 66.1 & 79.3 & - & 79.6 & 18.1 & - & 83.5 \\
TerraMind & ViT Large & \textbf{74.0} & 65.4 & 98.1 & - & 97.8 & 81.3 & 41.1 & 60.9 & 71.5 & 79.5 & 81.5 & - & 51.1 & 93.9 & 24.5 & 24.5 & 96.5 & 96.9 & 66.1 & - & - & - & - & - & 83.0 \\
\hline
{\model} v1 & ViT Nano & 66.8 & 61.5 & 98.2 & 50.3 & 95.7 & 39.5 & 35.4 & 53.0 & 60.6 & 78.8 & 82.5 & 82.5 & 61.1 & 96.0 & 20.4 & 19.7 & 97.4 & 97.4 & 75.6 & 74.8 & 70.2 & 75.5 & 17.1 & 72.0 & 82.1 \\
{\model} v1 & ViT Tiny & 69.6 & 63.5 & 98.7 & 53.2 & 97.9 & 72.5 & 38.5 & 60.3 & 71.5 & 79.7 & 85.0 & 85.5 & 60.6 & 97.7 & 19.8 & 19.2 & 97.6 & 97.7 & 78.2 & 76.4 & 74.4 & 76.9 & 15.8 & 73.5 & 85.2 \\
{\model} v1 & ViT Base & 72.0 & 68.6 & 98.2 & 51.2 & \cellcolor{colorone!25} \textbf{98.6} & 79.8 & 39.6 & 64.3 & \cellcolor{colorone!25} 77.8 & 79.8 & \cellcolor{colorone!25} \textbf{87.0} & \cellcolor{colorone!25} \textbf{86.0} & 62.4 & 97.1 & 18.5 & \cellcolor{colorone!25} \textbf{17.9} & \cellcolor{colorone!25} 97.6 & \cellcolor{colorone!25} \textbf{97.9} & \cellcolor{colorone!25} 81.8 & 82.2 & \cellcolor{colorone!25} 75.4 & 79.2 & 15.4 & 74.6 & 85.4 \\
{\model} v1 & ViT Large & 72.4 & 68.1 & 98.8 & 52.7 & 98.3 & 80.6 & 40.8 & \textbf{66.3} & \textbf{81.8} & 79.8 & 84.5 & - & 58.8 & \textbf{97.9} & 19.9 & 18.5 & 97.6 & 97.6 & 81.0 & - & - & - & - & - & 84.2 \\
\hline
{\model} v1.2 & ViT Nano & 68.0 & 60.8 & 98.1 & 53.7 & 96.5 & 54.0 & 36.3 & 52.7 & 58.1 & 78.9 & 75 & 75.5 & 61.5 & 95.4 & 20.4 & 19.9 & 96.3 & 96.6 & 79.2 & 78.1 & 71.1 & 75.1 & 16.8 & 72.4 & 80.1 \\
{\model} v1.2 & ViT Tiny & 71.3 & 67.4 & 98.8 & 56.8 & 98.1 & 87.8 & 41.7 & 63.7 & 67.7 & 79.4 & 83.5 & 84.5 & 59.3 & 97.8 & 19.5 & 18.8 & 97.2 & 97.5 & 80.7 & 81.1 & 75.2 & 76.9 & 14.3 & 73.8 & 86.6 \\
{\model} v1.2 & ViT Small & 72.0 & 66.5 & 99.0 & 56.5 & 98.2 & \textbf{95.8} & 44.0 & 64.7 & 69.7 & 79.8 & 83.0 & 82.0 & 62.0 & \textbf{97.9} & 19.4 & 19.0 & 97.5 & 97.7 & 78.3 & 80.2 & \textbf{75.6} & \textbf{81.9} & \textbf{14.2} & \textbf{75.9} & \textbf{87.5} \\
{\model} v1.2 & ViT Base & 73.0 & \cellcolor{colorone!25} \textbf{70.8} & \cellcolor{colorone!25} \textbf{99.1} & \cellcolor{colorone!25} 56.1 & 98.3 & \cellcolor{colorone!25} 82.5 & 43.5 & \cellcolor{colorone!25} 65.8 & 75.8 & \cellcolor{colorone!25} \textbf{80.5} & 83.0 & 85.0 & 60.2 & 97.1 & 19.1 & 18.3 & 97.5 & 97.6 & 81.6 & \cellcolor{colorone!25} \textbf{82.5} & 74.8 & 78.3 & \cellcolor{colorone!25} 14.5 & \cellcolor{colorone!25} \textbf{75.9} & 87.3 \\
\hline
{\model} v1.3 & ViT Base + P & \cellcolor{colorone!25} 73.3 & 66.5 & 99.0 & 55.3 & 98.0 & 80.5 & \cellcolor{colorone!25} \textbf{44.4} & 65.7 & 72.3 & 79.7 & 84.5 & 79.5 & \cellcolor{colorone!25} 62.9 & \cellcolor{colorone!25} 97.7 & \cellcolor{colorone!25} \textbf{18.4} & 18.5 & 97.5 & 97.7 & 78.7 & 80.7 & 71.9 & 79.0 & 16.2 & 70.6 & 84.8 \\
{\model} v1.3 & ViT Base & 72.7 & 68.4 & 99.0 & 52.7 & 98.5 & 81.0 & 42.7 & 64.1 & 74.6 & \cellcolor{colorone!25} \textbf{80.5} & 86.5 & 85.5 & \cellcolor{colorone!25} \textbf{65.6} & \cellcolor{colorone!25} 97.3 & 19.1 & 18.1 & 97.5 & 97.8 & 77.3 & 78.3 & 74.7 & \cellcolor{colorone!25} 79.5 & 14.6 & 75.7 & \cellcolor{colorone!25} \textbf{87.5} \\
\end{tabular}}
    \caption{Finetuning results on research benchmarks and partner tasks. Due to resource constraints, we do not fine-tune large models on all tasks.  We run v1.3 with the Attention Pooler (``+P'') and without. We \hl{highlight} which ViT Base model wins between v1, v1.2, and v1.3.}
    \label{tab:finetune}
\end{table*}

%% file: sec/5_platform.tex
\section{Creating \& Serving Global Scale Embeddings}
\label{sec:platform}

Computing embeddings over the whole Earth is expensive. Alongside the {\model} v1.3 weights we release a precomputed embedding dataset, readable directly from object storage: 128-dimensional int8 vectors at 10\,m resolution, one per pixel per reference year.

Like the evaluations in Section \ref{sec:embedding_results}, embeddings are computed using Sentinel-1, Sentinel-2 and Landsat-8/9. Each year is split into twelve 30-day periods ending on January 1 of the following year, and for each modality and period the scenes covering a window are composited into one mosaic by first valid pixel, least cloudy first for Sentinel-2 and Landsat, giving 12 timesteps per modality.

Embeddings are annual, each computed from one calendar year of imagery. A Zarr array's shape is fixed when it is created, so the store holds one slot per year from 2017 to 2025 from the start; each year is filled independently, and a year not yet computed takes no storage.
The coverage area is the union of three things: land; full-resolution shorelines, which
keep barrier islands and atolls; and water that all three modalities observe consistently,
which we take to be at least 11 months in at least 7 of the years 2017--2025 and which
selects most of the coastal zone. A window touching that area is then computed in full,
water included. The grid follows the UTM system, which spans $80^\circ$S to $84^\circ$N.
Within it, the coverage area runs from $59.5^\circ$S to $83.6^\circ$N and does not
include Antarctica.

Embeddings for 2025, 2024 and 2023 are global, and for 2017--2022 cover select countries and are in
progress over the same extent. We are still computing embeddings, so the coverage
described here is a snapshot; the dataset publishes a coverage layer of the extent each
year will reach when complete.

\subsection{Inference at scale}
\label{sec:platform_inference}

Inference is embarrassingly parallel over space and time, so the pipeline is organized
as a work queue. We tile each UTM zone, discard tiles that fall outside the zone or
touch none of the coverage area, and enqueue the rest onto a Beaker \cite{guerquin2022beaker} queue as
independent work units, each an $8192 \times 8192$ pixel block for one reference year.
A block is split into $2048 \times 2048$ windows, and every window that touches the
coverage area is computed in full, water included. A supervisor keeps the queue filled and the worker
pool at its target size; workers claim units, materialize the imagery they need, run
the model, and write results directly into the store.

The model runs at patch size 2 with 10\,m output (Section \ref{sec:model}). Each window
is split into $480 \times 480$ pixel crops, each a single forward pass; adjacent crops
overlap by 32 pixels and each keeps 16 of them at a shared edge. Workers run on NVIDIA
H100 nodes and complete a full unit in a median of 30 minutes, or roughly $0.5$
GPU-hours per unit. To add capacity we also run workers on NVIDIA A100 (40\,GB)
virtual machines on Google Cloud, which take units from the same queue and write to the
same store. A $480$-pixel crop does not fit in 40\,GB, so these workers use $384$-pixel
crops; on the same windows their output matches the H100 output at a mean cosine
similarity of $0.9999$. An A100 completes a full unit in a median of 84 minutes.
Throughput scales with the size of the pool. 

\input{sec/storage_layout}

Each work unit is idempotent: a completion marker is written only after its embeddings
land, and enumeration skips units already marked, so the build is interruptible and
resumable. Over a multi-day run workers are preempted and occasionally hang, and none of
that costs more than the units a worker holds at the time.


Materializing a unit's imagery is network-bound and takes minutes, so each worker
overlaps it with inference. The queue lets a worker hold two units at once: while the
GPU runs one, a separate process downloads the next unit's imagery to local disk, and
inference on that unit then reads from disk. Only download time that outlasts the
inference ahead of it adds to a unit's time. On H100 workers, a unit's time grows with
the number of windows it writes, at 1.9 minutes per window plus about 2 minutes per unit.

\subsection{Removing tiling artifacts using Searchlight} \label{sec:searchlight}

\begin{figure}[h]
\centering

\begin{minipage}{0.25\linewidth}
    \centering
    \includegraphics[width=\linewidth]{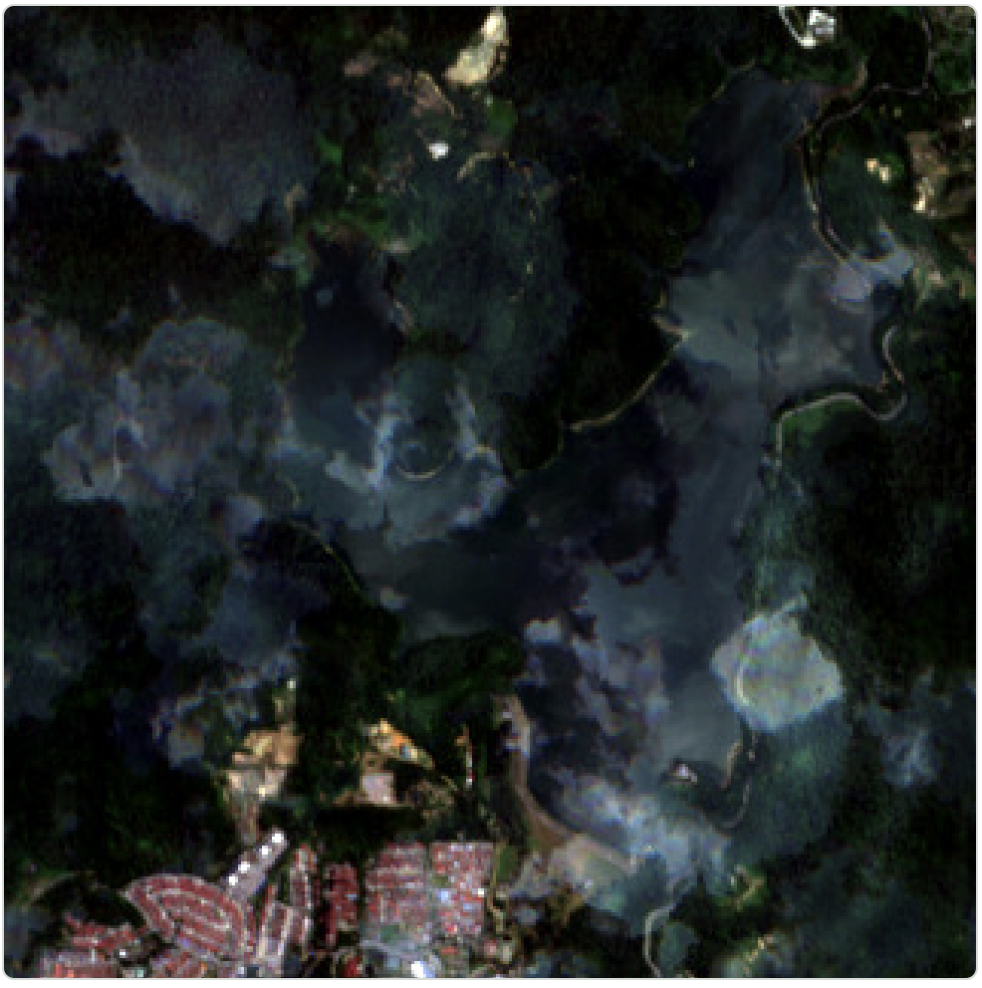}
    \caption*{S2}
\end{minipage}
\hspace{0.05\linewidth}
\begin{minipage}{0.25\linewidth}
    \centering
    \includegraphics[width=\linewidth]{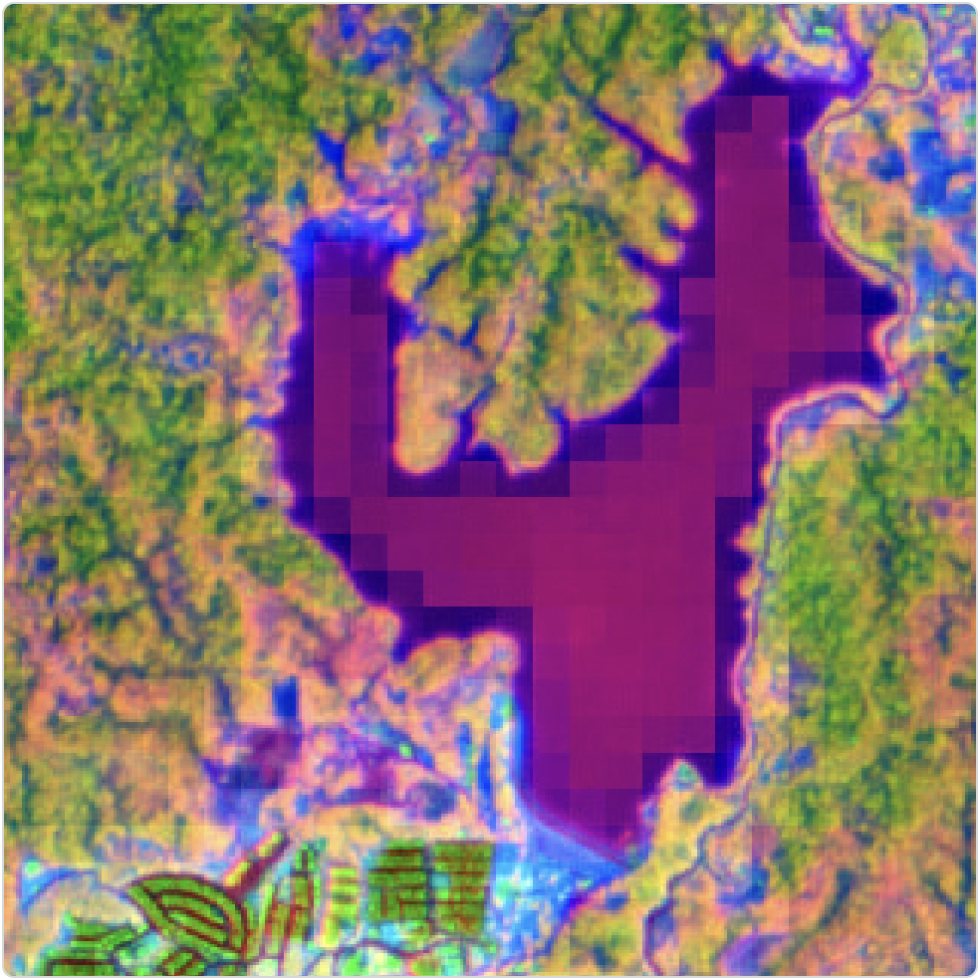}
    \caption*{Tiled Inference}
\end{minipage}
\hspace{0.05\linewidth}
\begin{minipage}{0.25\linewidth}
    \centering
    \includegraphics[width=\linewidth]{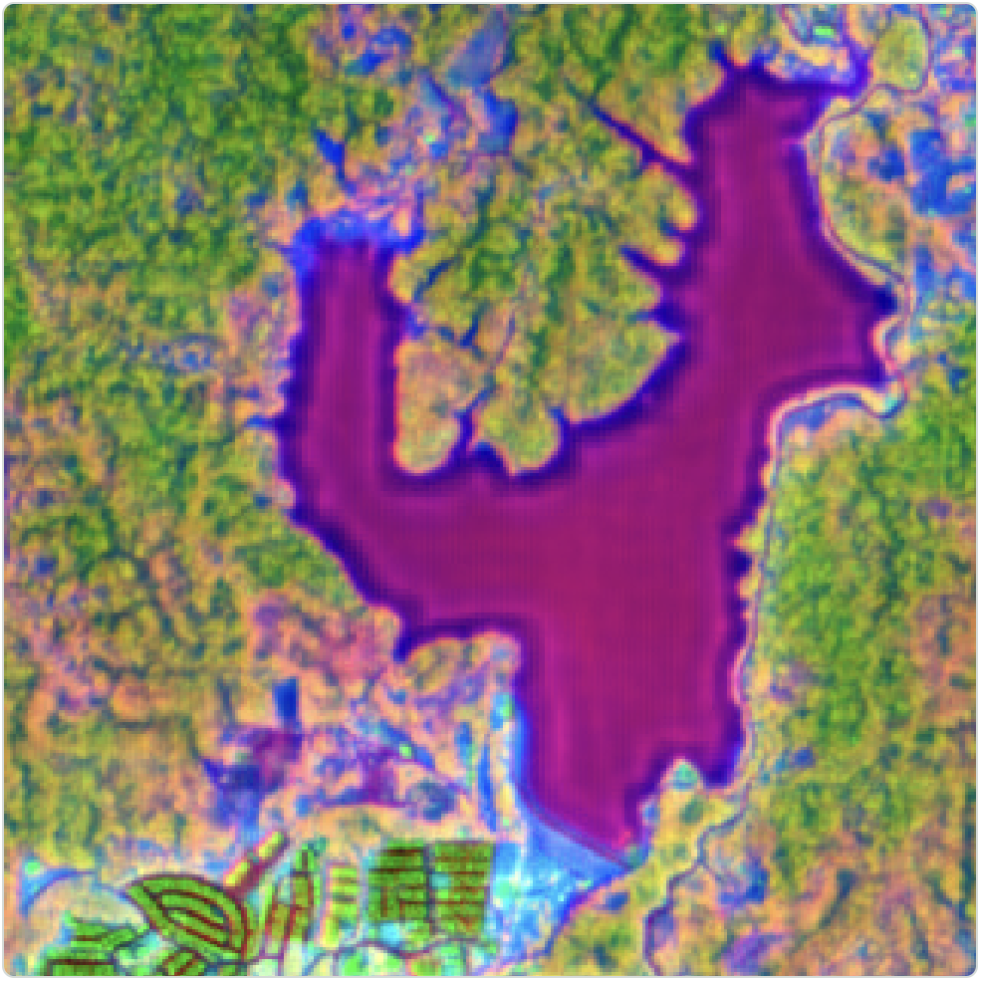}
    \caption*{Searchlight Inference}
\end{minipage}
\caption{Tiled inference -- even with significant overlap between crops -- introduces artifacts between crops, especially on homogeneous areas like water. These artifacts disappear when using searchlight inference.}
\label{fig:searchlight}
\end{figure}

A common challenge when producing maps from ViT-based models is tiling artifacts. Pixels at the edge of input crops have less spatial context, and that context is biased in a certain direction; this can create discrepancies between neighbouring pixels which were processed by different crops (Figure \ref{fig:searchlight}). 

To address this issue, we run inference using Searchlight, producing artifact-free embeddings. Searchlight inference uses neighborhood attention \cite{hassani2026generalized}, which lets it run on inputs much larger than those seen during training, since compute cost grows only linearly with input size. It also reduces tiling artifacts by making inference translation-invariant: apart from boundary effects, the model applies exactly the same operation at every pixel. Instead of passing batches of crops to the model, we instead pass a single, large ``domain'' to the model. We then apply an attention mask such that every pixel sees the $16\times16$ window centered around it, effectively mirroring the case where it is the center pixel of a $16 \times 16$ window.

Searchlight introduces attention (neighborhood vs full) and input size (large vs small) discrepancies between training and inference. To ensure this does not negatively affect embedding quality, we measure PASTIS performance using tiled inference and Searchlight inference. PASTIS validation mIoU remains stable (we obtained 60.4 using Searchlight and 59.8 using tiled inference).

\subsection{Seamless windows}
\label{sec:platform_padding}

\input{sec/window_padding}

Searchlight removes seams between crops, but each $2048 \times 2048$ window is still
run on its own, so pixels at its edge see context from one side only and a faint seam
remains every 20.48\,km (Figure \ref{fig:window_padding}). We therefore materialize
and run each window with a 48-pixel margin on every side and write only its central
$2048 \times 2048$ pixels, so shards and the write path are unchanged. The mean cosine
similarity between pixels adjacent across a window edge, relative to adjacent pixels
just inside it, rises from 0.960 to 0.994 (median over six edges at three sites), while
window interiors are unchanged (cosine similarity $\geq 0.9999$). A 112-pixel margin
gives no further gain, and 48 pixels adds 9.6\% to the pixels computed per window. 2025,
most of 2024 and a small part of 2019 and 2023 were built before the change, so each
block records its margin in its completion marker and provenance.

\subsection{Dataset layout}
\label{sec:platform_layout}

The embeddings are a Zarr v3 store with one array per UTM zone over $(\text{year},
\text{dimension}, \text{northing}, \text{easting})$. It uses the sharding codec, which
packs many chunks into a single storage object. Each zone is held in its northern CRS (EPSG:326NN)
with a northing axis continuing below zero south of the equator, so one array spans
the full UTM range, $80^\circ$S to $84^\circ$N, with coordinates monotonic across the equator (Figure
\ref{fig:storage_layout}). The conventional alternative pairs each northern CRS with a
southern one (EPSG:327NN) whose northing restarts at 10{,}000\,km at the equator, so any read crossing it would have to join two arrays.

\input{sec/matryoshka_bytes}

Shard size is pinned to the inference window at $2048 \times 2048$ pixels, which makes
the write path lock-free by construction: one window produces exactly one object, owned
by exactly one worker, so hundreds of writers proceed with no coordination and no
partial-object states.

Within a shard we chunk at $256 \times 256$ pixels under Zstandard level 3, chosen by
measuring bytes moved over the network across a sweep of layout variants. A chunk is the
smallest unit the store can transfer, and compressed chunks are not seekable, so a
single-pixel query still moves the whole chunk containing it, roughly 3\,MB. That is
cheap occasionally but wasteful in bulk, so the store is built for area reads.

Chunking also decides whether a nested embedding \cite{NEURIPS2022_c32319f4} is cheaper
to read and not just to store: a prefix saves bytes only when the dimensions beyond it
sit in a different chunk. We chunk the embedding axis at 64, the narrowest width the model is
trained to emit (Section \ref{sec:compaction}), so reading $[\ldots,{:}64]$ fetches one
chunk instead of two and moves 3.04\,MB instead of 6.05\,MB per tile (Figure
\ref{fig:matryoshka_bytes}).

The store follows the \texttt{geoemb} embeddings-zarr
convention,\footnote{\url{https://github.com/geo-embeddings/embeddings-zarr-convention}}
and records model identity, sources and quantization parameters as attributes, so any
byte is traceable to the code that produced it.

\subsection{Quantization}
\label{sec:platform_quantization}

Embeddings are stored as int8 using a signed-power scheme
\cite{brown2025alphaearthfoundationsembeddingfield}. Writing $s = 127.5$ for the scale
and $\gamma = \tfrac{1}{2}$ for the exponent, a float value $x$ is encoded and recovered
by
\begin{align}
  q &= \Big[\big\lfloor s\,\operatorname{sgn}(x)\,\lvert x \rvert^{\gamma}
       \big\rceil\Big]_{-127}^{\,127},
  &
  \hat{x} &= \operatorname{sgn}(q)\,\left\lvert \frac{q}{s} \right\rvert^{1/\gamma},
  \label{eq:quantization}
\end{align}
where $\lfloor\,\cdot\,\rceil$ rounds to the nearest integer and
$[\,\cdot\,]_{a}^{\,b}$ clamps to $[a,b]$; the value $-128$ is reserved as nodata.
We quantize the output as emitted rather than L2-normalizing it, which would discard
magnitude that downstream classifiers use; in practice nearly all coordinates of the
student head's LayerNorm output fall inside the scheme's $[-1,1]$ range, and the clamp
handles the rest. Median cosine similarity
between a float embedding and its dequantized form is $0.99997$.

\subsection{Provenance}
\label{sec:platform_provenance}

Each block writes a provenance record beside its completion marker. For every window
and modality it lists the scenes composited into each 30-day mosaic, in compositing order.
Each record also carries the window's materialized bounds and its margin. Each scene is
resolved to the public
source its imagery was read from: Sentinel-1 RTC and Sentinel-2 L2A from Microsoft
Planetary Computer, and Landsat Collection~2 from Google Cloud's public Landsat bucket, with
its ESA or USGS product name. We release these records as GeoParquet tables per
reference year: one row per window, one per scene in each 30-day mosaic, and one per
scene with its acquisition time and footprint. For 2025 that is about 450{,}000
windows, 21 million mosaic entries and 730{,}000 scenes, in 150\,MB.

\subsection{Distribution}
\label{sec:platform_distribution}

Hosting and serving the dataset is a significant challenge in its own right. At 10\,m,
128 int8 dimensions per pixel come to roughly 150\,TB per reference year at global
coverage, well over a petabyte across 2017 to 2025, and the dataset is meant to be
read directly from object storage by anyone, at any scale. We are very grateful to
Source Cooperative, which hosts the dataset and serves it openly to the
community.\footnote{\url{https://source.coop/olmoearth/olmoearth-v1-3-embeddings}}

The store is read directly through Source Cooperative's S3-compatible endpoint with any
Zarr client, with no API key or tile server needed. Worked examples ship with the dataset.

\subsection{Interactive exploration}
\label{sec:platform_explorer}

We also publish a browser application at
\url{https://olmoearth.allenai.org/embeddings}, which shows the false-color layer of
Section \ref{sec:platform_pca} as a web map and runs simple analyses on the embeddings
themselves, such as finding places similar to one the user clicks or clustering an area
they draw. Any result can be exported as a Python script that reproduces it against the
dataset.

\subsection{False-color visualization}
\label{sec:platform_pca}

Embeddings are not directly viewable, so the explorer shows a false-color rendering: a PCA
basis fitted on a stratified sample, projected to three components as a uint8 RGB
raster. Three components capture only a fifth to two fifths of local variance, so this
is a visualization and not a reduced-dimension version of the embeddings. It is not part of the released dataset. It lives in
its own store, reprojected from the embeddings' UTM grid to Web Mercator (EPSG:3857) on
the standard XYZ scheme, so a pyramid level is a web-map zoom level and a client reads a
bounded number of chunks at any extent.

%% file: sec/storage_layout.tex
\begin{figure}[t]
\centering
\definecolor{oecream}{HTML}{FAF2E9}
\begin{tikzpicture}[
  font=\scriptsize,
  frame/.style={draw=colordefault, line width=0.7pt},
  cell/.style={draw=colordefault!22, line width=0.15pt},
  hot/.style={fill=colorone, draw=colorone, line width=0.15pt},
  inert/.style={fill=oecream},
  hdr/.style={text=colordefault, font=\scriptsize\bfseries, align=center},
  sub/.style={text=colordefault!70, font=\tiny, align=center},
  flow/.style={-{Stealth[length=3.5pt,width=2.5pt]}, draw=colordefault!55,
               line width=0.5pt},
]

\def\Rg{0.92}
\def\Cx{1.65}
\def\Cy{0.75}
\fill[inert] (\Cx,\Cy) circle [radius=\Rg];
\foreach \lat in {-55,-27.5,0,27.5,55}{
  \pgfmathsetmacro{\ly}{\Cy+\Rg*sin(\lat)}
  \pgfmathsetmacro{\lrx}{\Rg*cos(\lat)}
  \pgfmathsetmacro{\lry}{0.21*\Rg*cos(\lat)}
  \draw[cell] (\Cx-\lrx,\ly)
    arc [start angle=180, end angle=540, x radius=\lrx, y radius=\lry];
}
\pgfmathsetmacro{\za}{\Rg*sin(30)}
\pgfmathsetmacro{\zb}{\Rg*sin(45)}
\fill[hot] (\Cx,\Cy+\Rg)
  arc [start angle=90, end angle=-90, x radius=\za, y radius=\Rg]
  arc [start angle=-90, end angle=90, x radius=\zb, y radius=\Rg] -- cycle;
\foreach \k in {-6,...,6}{
  \pgfmathsetmacro{\mx}{\Rg*sin(\k*15)}
  \draw[cell] (\Cx,\Cy+\Rg)
    arc [start angle=90, end angle=-90, x radius=\mx, y radius=\Rg];
}
\draw[frame] (\Cx,\Cy) circle [radius=\Rg];
\node[hdr, anchor=south] at (\Cx,2.10) {world};
\node[sub, anchor=north, text width=3.2cm] at (\Cx,-0.62)
  {60 UTM zones, $6^\circ$ of longitude each, $80^\circ$S to $84^\circ$N};

\draw[flow] (3.50,0.75) -- (4.10,0.75);

\fill[inert] (4.3,-0.35) rectangle (5.3,1.85);
\foreach \i in {1,...,3}{\draw[cell] (4.3+\i*0.25,-0.35) -- (4.3+\i*0.25,1.85);}
\foreach \j in {1,...,10}{\draw[cell] (4.3,-0.35+\j*0.2) -- (5.3,-0.35+\j*0.2);}
\fill[hot] (4.80,0.85) rectangle (5.05,1.05);
\draw[frame] (4.3,-0.35) rectangle (5.3,1.85);
\node[hdr, anchor=south, text width=2.2cm] at (4.8,2.10)
  {zone\\[-2pt]\mdseries$(T, D, H, W)$};
\node[sub, anchor=north, text width=2.2cm] at (4.8,-0.62)
  {$1{,}867{,}776 \times 98{,}304$ px, a $48 \times 912$ shard grid};

\draw[flow] (5.50,0.75) -- (6.10,0.75);

\fill[inert] (6.3,-0.45) rectangle (8.7,1.95);
\foreach \i in {1,...,7}{\draw[cell] (6.3+\i*0.3,-0.45) -- (6.3+\i*0.3,1.95);}
\foreach \j in {1,...,7}{\draw[cell] (6.3,-0.45+\j*0.3) -- (8.7,-0.45+\j*0.3);}
\fill[hot] (7.20,0.45) rectangle (7.50,0.75);
\draw[frame] (6.3,-0.45) rectangle (8.7,1.95);
\node[hdr, anchor=south, text width=2.6cm] at (7.5,2.10)
  {shard\\[-2pt]\mdseries$(1, D, 2048, 2048)$};
\node[sub, anchor=north, text width=2.5cm] at (7.5,-0.62)
  {one inference window, $20.48$\,km, one object, one writer};

\draw[flow] (8.90,0.75) -- (9.50,0.75);

\fill[inert] (10.25,0.80) rectangle (11.15,1.70);
\draw[frame] (10.25,0.80) rectangle (11.15,1.70);
\fill[hot] (9.85,0.40) rectangle (10.75,1.30);
\draw[frame] (9.85,0.40) rectangle (10.75,1.30);
\draw[cell] (10.75,0.40) -- (11.15,0.80);
\draw[cell] (10.75,1.30) -- (11.15,1.70);
\draw[cell] (9.85,1.30) -- (10.25,1.70);
\node[sub, anchor=west] at (11.25,1.52) {dims $64{:}128$};
\node[sub, anchor=west, text=colorone] at (11.25,0.58) {dims $0{:}64$};
\node[hdr, anchor=south, text width=2.8cm] at (10.5,2.10)
  {chunk\\[-2pt]\mdseries$(1, 64, 256, 256)$};
\node[sub, anchor=north, text width=3.0cm] at (10.5,-0.62)
  {the smallest unit transferred; $D{=}128$ spans two of them};

\end{tikzpicture}
\vspace{-2pt}
\caption{Layout of the {\model} embedding dataset. Each UTM zone is one Zarr array over
(reference year, embedding dimension, northing, easting) at 10\,m, sparse so unwritten
regions cost nothing. A \emph{shard} is the unit of storage and is pinned to one
inference window, so a window writes exactly one object and concurrent writers never
contend. A \emph{chunk} is the unit of transfer; chunking the embedding axis at 64 means
a Matryoshka read of the first 64 dimensions fetches one chunk rather than two.}
\label{fig:storage_layout}
\end{figure}

%% file: sec/window_padding.tex
\begin{figure}[h]
\centering
\begin{minipage}{0.3\linewidth}
    \centering
    \includegraphics[width=\linewidth]{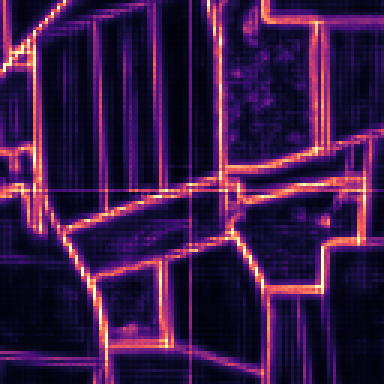}
    \caption*{No margin}
\end{minipage}
\hspace{0.05\linewidth}
\begin{minipage}{0.3\linewidth}
    \centering
    \includegraphics[width=\linewidth]{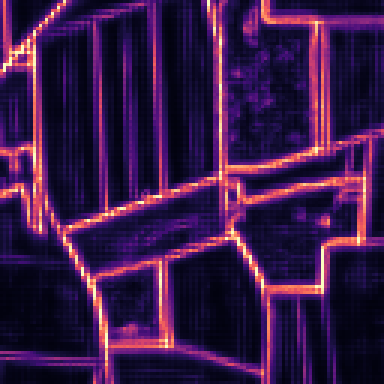}
    \caption*{48\,px margin}
\end{minipage}
\caption{Dissimilarity between adjacent pixels where four inference windows meet
(bright is a boundary). Without a margin the window edges show as a faint cross; with
a 48-pixel margin they disappear and field boundaries are unchanged.}
\label{fig:window_padding}
\end{figure}

%% file: sec/matryoshka_bytes.tex
\begin{wrapfigure}{r}{0.46\linewidth}
\vspace{-8pt}
\centering
\begin{tikzpicture}[
  font=\scriptsize,
  mbar/.style={draw=none},
  lbl/.style={text=colordefault, font=\scriptsize},
  val/.style={text=colordefault, font=\scriptsize\bfseries},
  sub/.style={text=colordefault!70, font=\tiny},
]
\def\sx{0.727}

\node[lbl, anchor=west] at (0,1.55) {all 128 dims, 2 chunks};
\fill[mbar, fill=colorone] (0,0.95) rectangle (6.05*\sx,1.35);
\node[val, anchor=west] at (6.05*\sx+0.1,1.15) {6.05\,MB};

\node[lbl, anchor=west] at (0,0.55) {first 64 dims, 1 chunk};
\fill[mbar, fill=colorone!45] (0,-0.05) rectangle (3.04*\sx,0.35);
\node[val, anchor=west] at (3.04*\sx+0.1,0.15) {3.04\,MB};

\draw[draw=colordefault!30, line width=0.4pt] (0,-0.2) -- (0,1.5);
\node[sub, anchor=west, align=left] at (0,-0.55)
  {measured on the wire; $4.19$\,MB\\per chunk once inflated};
\end{tikzpicture}
\vspace{-6pt}
\caption{Bytes transferred to read one $256\times256$ tile of embeddings. Because a
whole chunk is the smallest unit the store can move, reading the Matryoshka
$[\ldots,{:}64]$ prefix costs half of the full 128 dimensions rather than the same.}
\label{fig:matryoshka_bytes}
\vspace{-10pt}
\end{wrapfigure}
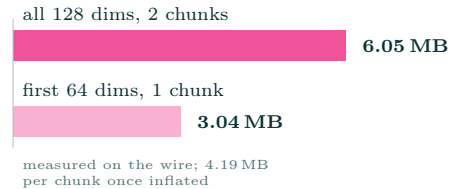

%% file: sec/5a_env_impact.tex
\section{Environmental Impact} \label{sec:env_impact}

\subsection{Pretraining}

We use the same technique as OlmoEarth v1 and OlmoEarth v1.2 to measure the pretraining cost of the OlmoEarth v1.3 model (Table \ref{tab:environmental-impact}). As with OlmoEarth v1 and OlmoEarth v1.2, this estimate represents a lower bound since it does not account for hardware manufacturing, transportation, etc. We significantly decreased the pretraining cost of OlmoEarth v1.3 compared to OlmoEarth v1.2 (and therefore OlmoEarth v1). Compared to OlmoEarth v1 / v1.2, pretraining OlmoEarth v1.3 Base required 18\% / 53\% of the GPU hours, and emitted 22\% / 50\% of the tCO$_2$eq. 

\input{sec/env_impact}

\subsection{Computing embeddings}

Computing global annualized embeddings dwarfs OlmoEarth v1.3's pretraining costs. One year of global embeddings costs approximately 14,000 H100 hours (as noted in Section \ref{sec:platform_inference}, our inference mixes H100s and A100s, so this is an estimate). This is $\approx 26\times$ the pretraining cost per year of annualized embeddings. 

We selected patch size 2 for inference as a balance between performance and compute cost (Section \ref{sec:embedding_ps}). Anyone generating embeddings locally who is limited by compute can produce 10m embeddings using an encoder patch size of 4 for roughly $5\times$ less compute, at the cost of slightly lower performance (PASTIS validation mIoU drops from 59.8 at patch size 2 to 58.6 at patch size 4).

%% file: sec/env_impact.tex
\begin{table}[htpb]
    \centering
\resizebox{3.25in}{!}{%
\begin{tabular}{l | c c c c c c}
 & &  & & Energy & Carbon & Water \\
Model & Size & Hardware & GPU Hrs & (kWh) & (tCO$_2$eq) & (kL) \\
\hline
{\model} v1 & Nano & H100 & 1,149 & 195 & 0.08 & 0.30 \\
{\model} v1 & Tiny & H100 & 1,149 & 205 & 0.08 & 0.32 \\
{\model} v1 & Base & H100 & 2,989 & 803 & 0.32 & 1.24 \\
\hline
{\model} v1.2 & Nano & H100 & 1,151 & 191 & 0.09 & 0.30 \\
{\model} v1.2 & Tiny & H100 & 1,132 & 225 & 0.10 & 0.35 \\
{\model} v1.2 & Small & H100 & 1,253 & 265 & 0.12 & 0.41 \\
{\model} v1.2 & Base & H100 & 1,012 & 317 & 0.14 & 0.49 \\
\hline
{\model} v1.3 & Base & H100 & 539 & 161 & 0.07 & 0.25 \\
\end{tabular}
}
\caption{Approximate environmental impact of pretraining {\model} v1, v1.2 and v1.3.
}
\label{tab:environmental-impact}
\end{table}

%% file: sec/6_discussion.tex
\section{Discussion}

Our mission with OlmoEarth is to make it easy to get high quality insights from Earth Observations. Embeddings play an important role in this mission, allowing practitioners to quickly and cheaply explore our changing planet. OlmoEarth v1.3 contributes in two ways; by halving pretraining costs, we hope more researchers can investigate geospatial foundation modelling, and by precomputing embeddings we hope it becomes easier for anyone to make high quality maps without running the model at all. 

We release annualized embeddings at a 10\,m resolution: global for 2023, 2024 \&--2025, and for select countries for 2017--2022.

Users can download either the full 128-dimensional embeddings, or the 64-dimensional prefix. Many tasks require more specific time ranges; for such tasks, we point users to OlmoEarth Studio (\url{https://olmoearth.allenai.org/platform}), which runs the same model over user-chosen time windows. OlmoEarth v1.3 can also be run locally to make embeddings independently on local compute.

We release OlmoEarth v1.3 and the embeddings under the same “OlmoEarth Artifact License” as OlmoEarth v1 and v1.2. This open license restricts use for military, defense-related, and extractive industry applications.

%% file: sec/appendix.tex
\appendix

\section{Full $k$NN / LP results}

Table \ref{tab:knnlp_subset} compares OlmoEarth-Base models to "best non-OlmoEarth models". We share the full results in Table \ref{tab:knnlp}.

\input{sec/knnlp}

%% file: sec/knnlp.tex
\sethlcolor{colorone!25}
\begin{table*}[htpb]
    \centering
    \resizebox{\textwidth}{!}{%


    \begin{tabular}{ll|cccccccccccccccccc|cccccc}

\hdr{} & \hdr{} & \hdr{m-bigearthnet} & \hdr{m-so2sat} & \hdr{m-brick-kiln} & \hdr{m-forestnet} & \hdr{m-eurosat} & \hdr{BreizhCrops} & \hdr{CropHarvest-PRC} & \hdr{CropHarvest-PRC} & \hdr{CropHarvest-PRC} & \hdr{CropHarvest-Togo} & \hdr{CropHarvest-Togo} & \hdr{CropHarvest-Togo} & \hdr{m-cashewplant} & \hdr{m-SA-crop-type} & \hdr{PASTIS} & \hdr{PASTIS} & \hdr{MADOS} & \hdr{Sen1Floods11} & \hdr{AWF} & \hdr{AWF} & \hdr{AWF} & \hdr{Nandi} & \hdr{Nandi} & \hdr{Nandi} \\
 & Modalities & S2 & S2 & S2 & L8 & S2 & S2 & S1 & S2 & S1,S2 & S1 & S2 & S1,S2 & S2 & S2 & S1 & S2 & S2 & S1 & L8 & S1 & S2 & L8 & S1 & S2 \\
 & Time series & \ccross & \ccross & \ccross & \ccross & \ccross & \ccheck & \ccheck & \ccheck & \ccheck & \ccheck & \ccheck & \ccheck & \ccross & \ccross & \ccheck & \ccheck & \ccross & \ccross & \ccheck & \ccheck & \ccheck & \ccheck & \ccheck & \ccheck \\
 & Method & kNN & kNN & kNN & kNN & kNN & LP & LP & LP & LP & LP & LP & LP & LP & LP & LP & LP & LP & LP & kNN & kNN & kNN & kNN & kNN & kNN \\
Model & Metric & $\mu \text{F1}$ & Acc. & Acc. & Acc. & Acc. & Acc. & Acc. & Acc. & Acc. & Acc. & Acc. & Acc. & mIOU & mIOU & mIOU & mIOU & mIOU & mIOU & Acc. & Acc. & Acc. & Acc. & Acc. & Acc. \\
\hline
Anysat \cite{astruc2024anysat} & ViT Base & 54.5 & 36.5 & 77.0 & 34.0 & 77.6 & 62.7 & 57.0 & 74.1 & 72.3 & 69.6 & 78.4 & 74.5 & 24.6 & 27.2 & 24.2 & 41.9 & 41.3 & 77.8 & 60.0 & 60.0 & 64.0 & 47.4 & 20.5 & 55.8 \\
Clay \cite{clayfoundation} & ViT Large & 48.8 & 38.7 & 92.8 & 40.3 & 86.1 & 57.0 & 56.7 & 66.5 & 63.4 & 78.4 & 67.0 & 67.3 & 30.8 & 23.1 & 19.9 & 22.6 & 47.5 & 78.9 & 59.5 & 61.5 & 56.5 & 52.2 & 23.4 & 57.0 \\
CopernicusFM \cite{wang2025towards} & ViT Base & 64.6 & 50.3 & 91.2 & - & 91.4 & 65.5 & 55.1 & 72.8 & 74.4 & 77.5 & 75.8 & 70.6 & 32.2 & 28.4 & 15.9 & 32.1 & 63.9 & 77.6 & - & 59.0 & 67.0 & - & 24.4 & 59.8 \\
CROMA \cite{fuller2024croma} & ViT Base & 61.3 & 51.3 & 94.0 & - & 93.1 & 69.0 & 56.8 & 74.3 & 75.1 & 76.8 & 80.7 & 76.5 & 24.9 & 30.3 & 26.3 & 44.7 & 60.4 & 78.8 & - & 67.5 & \textbf{79.0} & - & 24.6 & 68.2 \\
CROMA \cite{fuller2024croma} & ViT Large & 59.2 & 48.2 & 91.9 & - & 93.6 & 68.2 & 56.2 & 72.7 & 71.0 & 79.4 & 76.5 & 81.0 & 27.0 & 30.4 & 25.9 & 42.7 & 66.4 & 78.8 & - & 62.5 & 71.0 & - & 26.1 & 68.2 \\
DINOv3 \cite{siméoni2025dinov3} & ViT Base & 51.0 & 47.1 & 91.3 & 43.3 & 86.6 & 31.3 & - & 64.5 & - & - & 67.0 & - & 23.5 & 26.7 & - & 18.1 & 53.5 & - & 48.5 & - & 61.0 & 42.5 & - & 54.1 \\
DINOv3 \cite{siméoni2025dinov3} & ViT Large & 55.8 & 45.3 & 89.9 & 46.0 & 84.0 & 31.3 & - & 66.1 & - & - & 68.3 & - & 24.5 & 26.1 & - & 17.4 & 52.4 & - & 43.5 & - & 60.5 & 39.9 & - & 49.8 \\
DINOv3 \cite{siméoni2025dinov3} & ViT Huge+ & 57.0 & 45.7 & 88.2 & 46.9 & 86.1 & 31.3 & - & 68.7 & - & - & 68.3 & - & 25.1 & 26.7 & - & 17.4 & 48.1 & - & 47.5 & - & 54.0 & 43.5 & - & 55.4 \\
DINOv3 \cite{siméoni2025dinov3} & ViT 7B & 60.8 & 46.6 & 91.3 & 48.0 & 85.2 & 31.3 & - & 68.7 & - & - & 70.9 & - & 34.3 & 27.9 & - & 21.1 & 52.2 & - & 47.5 & - & 57.0 & 44.7 & - & 56.7 \\
DINOv3 Sat \cite{siméoni2025dinov3} & ViT Large & 60.2 & 44.0 & 91.4 & 44.2 & 89.2 & 31.3 & - & 70.1 & - & - & 68.6 & - & 32.4 & 28.5 & - & 22.5 & 57.5 & - & 42.5 & - & 69.5 & 35.5 & - & 48.4 \\
DINOv3 Sat \cite{siméoni2025dinov3} & ViT 7B & 61.6 & 50.1 & 91.4 & 47.0 & 91.3 & 31.3 & - & 72.2 & - & - & 71.9 & - & \textbf{54.1} & 31.7 & - & 26.3 & 59.7 & - & 49.5 & - & 68.5 & 31.8 & - & 42.5 \\
Galileo \cite{tseng2025galileo} & ViT Nano & 55.0 & 53.7 & 91.6 & - & 91.5 & 66.3 & 60.9 & 74.4 & 72.4 & 75.2 & 70.3 & 78.1 & 21.2 & 19.5 & 19.2 & 19.1 & 53.1 & 78.6 & - & 65.5 & 67.5 & - & 24.4 & 64.2 \\
Galileo \cite{tseng2025galileo} & ViT Tiny & 55.8 & 53.1 & 90.7 & - & 93.1 & 66.7 & 55.7 & 80.3 & 79.3 & 69.9 & 78.8 & 77.1 & 23.6 & 21.5 & 23.4 & 27.7 & 61.1 & 78.6 & - & 65.5 & 71.0 & - & 25.0 & 66.7 \\
Galileo \cite{tseng2025galileo} & ViT Base & 58.3 & 55.7 & 88.5 & - & 94.3 & 69.7 & 60.9 & \textbf{81.9} & 79.3 & 67.3 & 80.1 & 77.5 & 28.9 & 25.3 & 28.0 & 39.6 & \textbf{68.4} & 79.4 & - & 66.5 & 72.5 & - & 26.8 & 67.3 \\
Panopticon \cite{waldmann_shah_2025_panopticon} & ViT Base & 64.9 & 60.5 & 94.1 & \textbf{52.3} & 95.0 & 57.7 & 55.9 & 75.9 & 75.6 & 72.2 & 72.9 & 76.5 & 32.7 & 27.3 & 23.7 & 30.2 & 66.1 & 78.0 & 66.0 & 65.0 & 71.5 & 60.4 & 22.9 & 65.2 \\
Presto \cite{tseng2023lightweight} & ViT Nano & - & - & - & - & - & 60.9 & 59.3 & 74.3 & 76.6 & 78.4 & 81.4 & 72.5 & - & - & 16.3 & 28.2 & - & - & - & 60.5 & 53.5 & - & 25.2 & 57.6 \\
Prithvi v2 \cite{szwarcman2024prithvi} & ViT Large & 53.8 & 42.6 & 90.3 & 37.9 & 82.2 & 66.2 & - & 72.0 & - & - & 71.9 & - & 51.1 & 27.3 & - & 37.2 & 55.6 & - & 57.0 & - & 55.5 & 57.7 & - & 59.3 \\
Prithvi v2 \cite{szwarcman2024prithvi} & ViT Huge & 52.1 & 39.5 & 89.9 & 41.4 & 81.2 & 66.1 & - & 71.7 & - & - & 69.9 & - & 49.2 & 29.7 & - & 37.5 & 58.2 & - & 59.0 & - & 55.0 & 57.9 & - & 55.9 \\
Satlas \cite{bastani2023satlaspretrain} & Swin Base & 52.3 & 44.5 & 91.9 & 36.9 & 89.1 & 64.6 & 57.4 & 71.3 & - & 76.8 & 75.8 & - & 30.6 & 24.0 & 10.5 & 14.4 & 30.2 & 72.9 & 57.5 & 52.0 & 62.0 & 61.5 & 25.1 & 60.2 \\
TerraMind \cite{jakubik2025terramindlargescalegenerativemultimodality} & ViT Base & 63.9 & 46.7 & 92.9 & - & 91.4 & 66.4 & 57.0 & 75.1 & 75.6 & 74.2 & 74.8 & 77.5 & 46.0 & 30.4 & 22.7 & 40.9 & 66.0 & 78.7 & - & 66.0 & 69.5 & - & 24.7 & 64.2 \\
TerraMind \cite{jakubik2025terramindlargescalegenerativemultimodality} & ViT Large & 63.9 & 47.4 & 92.9 & - & 90.9 & 68.2 & 56.2 & 74.7 & 72.0 & 75.2 & 77.8 & 75.2 & 50.4 & 31.2 & 22.3 & 41.3 & 67.5 & 78.4 & - & 62.0 & 67.0 & - & 23.3 & 65.3 \\
\hline
{\model} v1 & ViT Nano & 59.5 & 54.3 & 94.7 & 38.8 & 91.7 & 64.1 & 58.0 & 79.5 & 74.3 & 73.5 & 81.7 & \textbf{83.7} & 25.5 & 23.6 & 18.1 & 35.0 & 55.2 & 78.2 & 73.0 & 61.5 & 69.5 & 57.7 & 24.8 & 67.4 \\
{\model} v1 & ViT Tiny & 59.4 & 61.8 & 93.3 & 40.5 & 93.7 & 64.0 & 58.3 & 78.6 & \textbf{80.3} & 75.8 & 85.6 & 82.4 & 24.7 & 23.2 & 21.4 & 40.1 & 58.6 & 78.5 & 75.5 & 64.0 & 76.0 & 60.7 & 24.7 & 69.0 \\
{\model} v1 & ViT Base & 62.4 & \cellcolor{colorone!25} 67.7 & 94.3 & 41.9 & \cellcolor{colorone!25} 95.5 & 70.9 & 56.8 & 73.4 & 75.4 & \cellcolor{colorone!25} \textbf{80.1} & \cellcolor{colorone!25} \textbf{87.3} & \cellcolor{colorone!25} 82.0 & \cellcolor{colorone!25} 32.3 & 28.9 & 29.7 & 50.6 & \cellcolor{colorone!25} 67.2 & 79.2 & \cellcolor{colorone!25} \textbf{77.0} & \cellcolor{colorone!25} 68.5 & \cellcolor{colorone!25} 77.5 & \cellcolor{colorone!25} \textbf{67.9} & 26.5 & 74.7 \\
{\model} v1 & ViT Large & 62.0 & \textbf{68.2} & 96.5 & 41.6 & \textbf{96.4} & 70.7 & 56.6 & 74.1 & 76.1 & 67.6 & 78.1 & 79.7 & 30.9 & 28.5 & 30.6 & 51.8 & 66.4 & \textbf{79.8} & 76.0 & 66.5 & 73.0 & 66.4 & 26.2 & 73.6 \\
\hline
{\model} v1.2 & ViT Nano & 59.5 & 55.3 & 94.8 & 38.5 & 91.4 & 67.5 & \textbf{63.7} & 77.5 & 80.2 & 73.9 & 79.4 & 82.7 & 26.8 & 22.4 & 22.5 & 35.9 & 58.1 & 78.7 & 71.5 & 64.0 & 71.5 & 66.9 & 25.9 & 71.0 \\
{\model} v1.2 & ViT Tiny & 61.7 & 58.8 & 92.6 & 39.9 & 93.8 & 70.6 & 59.2 & 77.6 & 80.0 & 74.5 & 82.0 & 79.4 & 27.1 & 24.4 & 24.9 & 45.3 & 63.0 & 78.7 & 74.0 & 64.5 & 76.0 & \textbf{67.9} & 28.4 & 67.3 \\
{\model} v1.2 & ViT Small & 64.8 & 65.7 & 91.8 & 44.4 & 94.5 & \textbf{79.1} & 59.0 & 79.1 & 78.1 & 73.5 & 83.0 & 80.4 & 28.5 & 26.7 & 28.3 & 50.7 & 61.3 & 79.3 & \textbf{77.0} & \textbf{69.0} & 74.5 & 67.2 & \textbf{29.7} & 73.2 \\
{\model} v1.2 & ViT Base & 65.5 & 67.3 & 94.7 & \cellcolor{colorone!25} 45.4 & 95.0 & \cellcolor{colorone!25} 73.3 & 58.2 & 76.7 & 77.7 & 74.8 & 81.4 & 81.0 & 29.6 & 32.0 & 31.1 & 54.2 & \cellcolor{colorone!25} 67.2 & 78.9 & 74.5 & 68.0 & 75.0 & 65.7 & 28.1 & 73.9 \\
\hline
{\model} v1.3 & ViT Base + P & \cellcolor{colorone!25} 65.9 & 55.4 & \cellcolor{colorone!25} 95.3 & 37.3 & 94.1 & 70.9 & 57.1 & 76.5 & 76.9 & 74.5 & 77.1 & 79.7 & 28.9 & 27.7 & \cellcolor{colorone!25} 31.4 & \cellcolor{colorone!25} \textbf{57.5} & 51.7 & \cellcolor{colorone!25} 79.5 & 73.0 & 65.0 & 76.0 & 66.4 & 27.0 & 71.5 \\
{\model} v1.3 & ViT Base & \cellcolor{colorone!25} \textbf{66.1} & 58.2 & \cellcolor{colorone!25} \textbf{96.8} & 41.7 & 93.4 & 73.2 & \cellcolor{colorone!25} 62.4 & \cellcolor{colorone!25} 77.7 & \cellcolor{colorone!25} 78.3 & 72.9 & 82.0 & 80.1 & 29.1 & \cellcolor{colorone!25} \textbf{32.8} & \cellcolor{colorone!25} \textbf{32.6} & \cellcolor{colorone!25} 54.4 & 67.1 & \cellcolor{colorone!25} 79.3 & 72.0 & 67.0 & 76.0 & 67.4 & \cellcolor{colorone!25} 28.5 & \cellcolor{colorone!25} \textbf{77.6} \\
\end{tabular}}
    \caption{kNN/Linear probe results on research benchmarks and real-world partner tasks. Not all models can run on all tasks due to incompatible input modalities. We run v1.3 with Attention Pooling (``+P'') and without. We \hl{highlight} which ViT Base model wins between v1, v1.2, and v1.3.}
    \label{tab:knnlp}
\end{table*}